\documentclass{article}

\usepackage[nonatbib, preprint]{neurips_2025}

\usepackage[utf8]{inputenc} 
\usepackage[T1]{fontenc}    
\usepackage{url}            
\usepackage{booktabs}       
\usepackage{amsfonts}       
\usepackage{nicefrac}       
\usepackage{microtype}      
\usepackage{xcolor}         
\usepackage{url}
\usepackage{cite}
\usepackage{graphicx} 
\usepackage{float}
\usepackage{booktabs}
\usepackage{times}
\usepackage{epsfig}
\usepackage{graphicx}
\usepackage{amsmath}
\usepackage{amsfonts}
\usepackage{cancel}
\usepackage{bbm}
\usepackage{mathtools}
\usepackage{amssymb}
\usepackage{pifont}
\usepackage{utfsym}
\usepackage{booktabs}
\usepackage{arydshln}
\usepackage{color}
\usepackage{colortbl}
\usepackage{subcaption}
\usepackage{multirow}
\usepackage{float}
\usepackage{rotfloat}
\usepackage{capt-of}
\usepackage{longtable}
\usepackage{diagbox}
\usepackage{makecell}
\usepackage{pgfplots}
\usepackage{algorithm}
\usepackage{xspace}
\usepackage{wrapfig}
\usepackage{enumitem}
\usepackage{fontawesome}
\usepackage{epigraph}
\usepackage{multirow}
\usepackage{multicol}
\usepackage{rotating}
\usepackage{amssymb}
\usepackage{pifont}
\usepackage{algorithm}
\usepackage{listings}
\usepackage{etoolbox}
\makeatletter
\AfterEndEnvironment{algorithm}{\let\@algcomment\relax}
\AtEndEnvironment{algorithm}{\kern2pt\hrule\relax\vskip3pt\@algcomment}
\let\@algcomment\relax
\newcommand\algcomment[1]{\def\@algcomment{\footnotesize#1}}
\renewcommand\fs@ruled{\def\@fs@cfont{\bfseries}\let\@fs@capt\floatc@ruled
  \def\@fs@pre{\hrule height.8pt depth0pt \kern2pt}%
  \def\@fs@post{}%
  \def\@fs@mid{\kern2pt\hrule\kern2pt}%
  \let\@fs@iftopcapt\iftrue}
\makeatother
\usepackage{algpseudocode}
\usepackage{colortbl}
\definecolor{mycolor_blue}{HTML}{E7EFFA}
\definecolor{mycolor_green}{HTML}{E6F8E0}
\definecolor{mycolor_gray}{HTML}{ECECEC}
\definecolor{pearDark}{HTML}{2980B9}

\usepackage{tcolorbox}
\usepackage{fbox}
\definecolor{textcolor1}{rgb}{0.25,0.5,0.5}
\definecolor{textcolor2}{rgb}{0.7,0.25,0.25}
\definecolor{linkc}{rgb}{0, 0.44, 0.74}
\definecolor{eqc}{rgb}{1, 0, 0}
\definecolor{myy}{RGB}{126,95,0}
\definecolor{mygray}{gray}{.9}
\definecolor{bblue}{RGB}{30,80,120}
\definecolor{mygray1}{gray}{.7}
\definecolor{ggray}{RGB}{127,127,127}
\definecolor{mygreen}{RGB}{93,174,86}
\definecolor{citecolor}{HTML}{229954}
\definecolor{light_green}{HTML}{F5FFFA}
\newcommand{\g}[1]{\textcolor{gray}{#1}}
\definecolor{scolor}{RGB}{111,168,220}
\definecolor{hcolor}{RGB}{111,176,81}
\definecolor{ocolor}{RGB}{224,103,102}
\definecolor{wcolor}{RGB}{246,178,107}
\newcommand{\SHOW}{\textcolor{scolor}{S}\textcolor{hcolor}{h}\textcolor{ocolor}{o}\textcolor{wcolor}{w}}
\usepackage[pagebackref=false,breaklinks=true,colorlinks=True,urlcolor=eqc,citecolor=citecolor,linkcolor=eqc,bookmarks=false,urlcolor=scolor]{hyperref}
\title{\SHOW-o2: Improved Native Unified Multimodal Models}

\author{%
	Jinheng Xie$^{1}$~
    \textbf{
        Zhenheng Yang$^2$~
	Mike Zheng Shou$^{1*}$}
	\\\\
	$^1$ Show Lab, National University of Singapore\quad
	$^2$ ByteDance
}

\begin{document}
\def\thefootnote{$*$ }\footnotetext{Corresponding Author } 
\maketitle

\begin{abstract}
  This paper presents improved native unified multimodal models, \emph{i.e.,} Show-o2, that leverage autoregressive modeling and flow matching. Built upon a 3D causal variational autoencoder space, unified visual representations are constructed through a dual-path of spatial (-temporal) fusion, enabling scalability across image and video modalities while ensuring effective multimodal understanding and generation. Based on a language model, autoregressive modeling and flow matching are natively applied to the language head and flow head, respectively, to facilitate text token prediction and image/video generation. A two-stage training recipe is designed to effectively learn and scale to larger models. The resulting Show-o2 models demonstrate versatility in handling a wide range of multimodal understanding and generation tasks across diverse modalities, including text, images, and videos. Code and models are released at  \href{https://github.com/showlab/Show-o}{https://github.com/showlab/Show-o}.
\end{abstract}

\section{Introduction}

Large language models (LLMs)~\cite{llama,qwen2.5} have achieved unprecedented performance levels, fueled by extensive web-scale text resources, substantial computational power, and billions of parameters. In the multimodal domain, large multimodal models (LMMs)~\cite{Qwen2.5-VL,llavaonevision,internvl} and visual generative models~\cite{sd3,sana,lumina2}, have also demonstrated exceptional capabilities in tasks such as general-purpose visual question answering and text-to-image/video generation. Given their success, unified multimodal models (UMMs)~\cite{wu2023next, team2024chameleon, showo} have been investigated to unify multimodal understanding and generation within a single model or system. In addition to multimodal understanding capability, this line of approaches seeks to simultaneously cultivate multimodal understanding and generation abilities in the model/system through pre-training, fine-tuning, or connecting tailored models. 

Here, we provide a comparative analysis of selected UMMs in Table~\ref{tab:model_compar}, focusing on two perspectives, including i) visual representations for understanding and generation and ii) the type of unified modeling. Generally, there are two approaches to incorporating visual representations for multimodal understanding and generation: i) a unified representation for both understanding and generation, as seen in works like Chameleon~\cite{team2024chameleon}, Transfusion~\cite{zhou2025transfusion}, and Show-o~\cite{showo}; and ii) decoupled representations, utilizing CLIP~\cite{clip} for multimodal understanding and variational autoencoder (VAE) for visual generation. To involve both multimodal understanding and generation capabilities, two primary methods have been explored: i) natively applying multimodal understanding and generation objectives within a single model and ii) tuning adapters to assemble tailored models. We refer the first type as \textit{native unified multimodal models}, distinguishing it from the second type that assembles tailored models. These principles, combined with autoregressive or diffusion modeling or both, contribute to the development of unified multimodal models.

Compared to existing UMMs that primarily focus on text and image, our approach explores model designs that provide substantial potential and scalability in natively unifying text, image, and video modalities. An overview of our approach is presented in Fig.~\ref{fig:overview}. Specifically, for visual inputs, we operate within the 3D causal VAE~\cite{wan2025} space, which is capable of accommodating both images and videos. Recognizing the distinct feature dependencies between multimodal understanding and generation, we construct unified visual representations that simultaneously capture rich semantic information and low-level features with intrinsic structures and textual details from the visual latents. This is achieved through a dual-path mechanism consisting of semantic layers, a projector, and a spatial (-temporal) fusion process. As the fusion process occurs within the 3D causal VAE space, when it comes to videos, semantic and low-level features are temporally aligned and fused with full-frame video information. 

Text embeddings and unified visual representations are structured into a sequence to go through a pre-trained language model and are modeled by a specific language head and flow head, respectively. Specifically, autoregressive modeling with causal attention is performed on the language head when dealing with text token prediction, and flow matching with full attention is applied to the flow head for image/video generation. Since the base language model lacks visual generation capabilities, we propose a two-stage training recipe to effectively learn such an ability while retaining the language knowledge, without requiring a massive text corpus. In the first stage, we mainly focus on pre-training the flow head for visual generation using (interleaved) text, image, and video data. In the second stage, the full model is fine-tuned with high-quality multimodal understanding and generation data. 

Extensive experimental results have demonstrated that our model surpasses the existing methods in terms of most metrics across multimodal understanding and visual generation benchmarks. Collectively, the main contributions of this paper can be summarized as:
\begin{itemize}
    \item We present an improved native unified multimodal model that seamlessly integrates autoregressive modeling and flow matching, enabling a wide range of multimodal understanding and generation across (interleaved) text, images, and videos.
    
    \item Based on the 3D causal VAE space, we construct unified visual representations scalable to both multimodal understanding and generation, image and video modalities by combining semantic and low-level features through a dual-path of spatial (-temporal) fusion mechanism.
    
    \item We design a two-stage training pipeline that effectively and efficiently learns unified multimodal models, retaining language knowledge and enabling effective scaling up to larger models, without requiring a massive text corpus.
    
    \item The proposed model demonstrates state-of-the-art performance on multimodal understanding and visual generation benchmarks, surpassing existing methods across various metrics.
    
\end{itemize}

\begin{table}[t]
\centering
\caption{Comparative analysis of selected unified multimodal models based on the type of visual representations and unified modeling for multimodal understanding and generation. In this context, \textbf{native und. \& gen.} refers to the direct decoding of output sequences into texts, images, and videos, as opposed to serving as conditions for decoding using external pre-trained decoders like Stable Diffusion. $^{*}$ indicates the method adopts two distinct models for multimodal understanding and generation, respectively. Diff. means the diffusion modeling. \textit{Please refer to the complete table in the appendix.}}
\label{tab:model_compar}
\resizebox{\linewidth}{!}{ 
\begin{tabular}{lcccccccc}
        \toprule
            \multirow{2}{*}{Methods} & \multicolumn{3}{c}{Und. \& Gen. Representation} & \multicolumn{3}{c}{Type of Unified Modeling} \\
            \cmidrule(lr){2-4}  \cmidrule(lr){5-7}
            & Unified & Decoupled & Support Video & Native Und. \& Gen. & Assembling Tailored Models & Paradigm \\
    	  \midrule
             Chameleon~\cite{team2024chameleon} & \checkmark & & \usym{2613} & \checkmark &  & AR \\
            Transfusion~\cite{zhou2025transfusion} & \checkmark & & \usym{2613} & \checkmark & & AR + Diff.\\
            Show-o~\cite{showo} & \checkmark & & \usym{2613} & \checkmark & & AR + Diff.\\
            VILA-U~\cite{vila-u} & \checkmark & & \checkmark & \checkmark & & AR \\
            Emu3~\cite{wang2024emu3} & \checkmark & & \checkmark & \checkmark & & AR \\
            LlamaFusion~\cite{lmfusion} & \checkmark & & \usym{2613}  & \checkmark &  & AR + Diff.\\
            \rowcolor{light_green}
            Show-o2 (Ours)& \checkmark & & \checkmark & \checkmark & & AR + Diff. \\
            \midrule
            Janus-Series~\cite{janus,ma2024janusflow,chen2025janus} & & \checkmark & \usym{2613} & \checkmark & & AR (+Diff)\\
            
            UnidFluid~\cite{unifluid} & & \checkmark & \usym{2613} & \checkmark & & AR + MAR \\ 
            Mogao~\cite{mogao} & & \checkmark & \usym{2613} & \checkmark & & AR + Diff. \\ 
            BAGEL~\cite{bagel} & & \checkmark & \checkmark & \checkmark & & AR + Diff. \\ 
            \midrule
            NExT-GPT~\cite{wu2023next} & & \checkmark & \checkmark  & & \checkmark & AR + Diff.\\
            SEED-X~\cite{seed-x} & & \checkmark & \usym{2613} & & \checkmark & AR + Diff.\\
            ILLUME~\cite{ILLUME} &  & \checkmark & \usym{2613} &  & \checkmark & AR + Diff. \\
            MetaMorph~\cite{tong2024metamorph} &  & \checkmark & \usym{2613}&  &  \checkmark & AR + Diff. \\
            MetaQueries~\cite{metaqueries} &  & \checkmark & \usym{2613}&  &  \checkmark & AR + Diff. \\
            \midrule
            TokenFlow$^{*}$~\cite{qu2024tokenflow} & \checkmark & & \usym{2613} & & \checkmark & AR\\
        \bottomrule

\end{tabular}
}
\vspace{-5mm}
\end{table}

\section{Related Work}
\subsection{Large Multimodal Models}
Building upon the advancements of large language models (LLMs)\cite{llama, qwen2.5}, large multimodal models (LMMs)\cite{llava, internvl, llavaonevision, Qwen2.5-VL} have showcased remarkable capabilities in general-purpose visual question answering. These approaches typically leverage pre-trained vision encoders to project visual features and align them within the embedding space of LLMs. Meanwhile, a growing number of encoder-free LMMs~\cite{showo, diao2024EVE, diao2025EVEv2} aim to directly align raw visual features within the LLM embedding space. However, these encoder-free methods often fall behind models that utilize image-text-aligned visual features in terms of performance. Beyond model architecture, recent studies~\cite{sharegpt4v, tong2024cambrian1, llavaonevision} have highlighted the critical role of high-quality instructional data in enhancing multimodal capabilities.

\subsection{Visual Generative Models}
Two prominent paradigms for visual generation, namely diffusion~\cite{rombach2022high, peebles2023scalable, uvit, pixart, Xie_2023_ICCV, wu2023tune, flow, sana, lumina2, show1, seawead2025seaweed} and autoregressive modeling~\cite{llamagen, kondratyuk2023videopoet, chen2020generative, randar, ARPG}, have been extensively studied in image and video generation in recent years. Diffusion-based methods typically employ optimized architectures that integrate pre-trained text encoders with denoising networks. In contrast, autoregressive methods often utilize LLM-based architectures and are trained through next-token prediction. Recently, several studies~\cite{mar, mardini, unifluid} have explored hybrid approaches that combine diffusion and autoregressive modeling to further advance visual generation capabilities.

\subsection{Unified Multimodal Models}

Building on the success of large multimodal and visual generative models, pioneering unified multimodal models (UMMs) such as Chameleon~\cite{team2024chameleon}, Show-o~\cite{showo}, and Transfusion~\cite{zhou2025transfusion} aim to integrate these capabilities into a single model through autoregressive or diffusion modeling or both. Further advancements~\cite{wang2024emu3, vila-u, unitok,unitoken,SemHiTok,dualtoken} have focused on optimizing the training pipeline and enhancing the semantics of discrete tokens, leading to improved performance. We refer to these approaches as \textit{native unified multimodal models}, as they inherently combine multimodal understanding and generation objectives within a unified architecture.

An alternative and promising direction~\cite{CoDI,uio2,dreamllm,seed-x,tong2024metamorph,metaqueries,blip3} for unifying multimodal understanding and generation involves assembling off-the-shelf specialized LMMs and visual generative models by tuning adapters or learnable tokens. Representative works~\cite{wu2023next,seed-x,metaqueries,blip3} have demonstrated the promising capabilities and intriguing properties of such assembled unified frameworks, highlighting their potential for further exploration.

\section{Methodology}
In this section, we introduce the overall framework (Section~\ref{sec:overall_framework}), which consists of two key components: i) the design of unified visual representations for multimodal understanding and generation, applicable to both images and videos, and ii) the native learning of multimodal understanding and generation capabilities. Subsequently, we present a two-stage training recipe (Section~\ref{sec:training_strategy}), which is designed to progressively learn and effectively scale up the unified multimodal model.

\subsection{Overall Framework}
\label{sec:overall_framework}
\begin{figure}[t]
    \centering
    \includegraphics[width=1\linewidth]{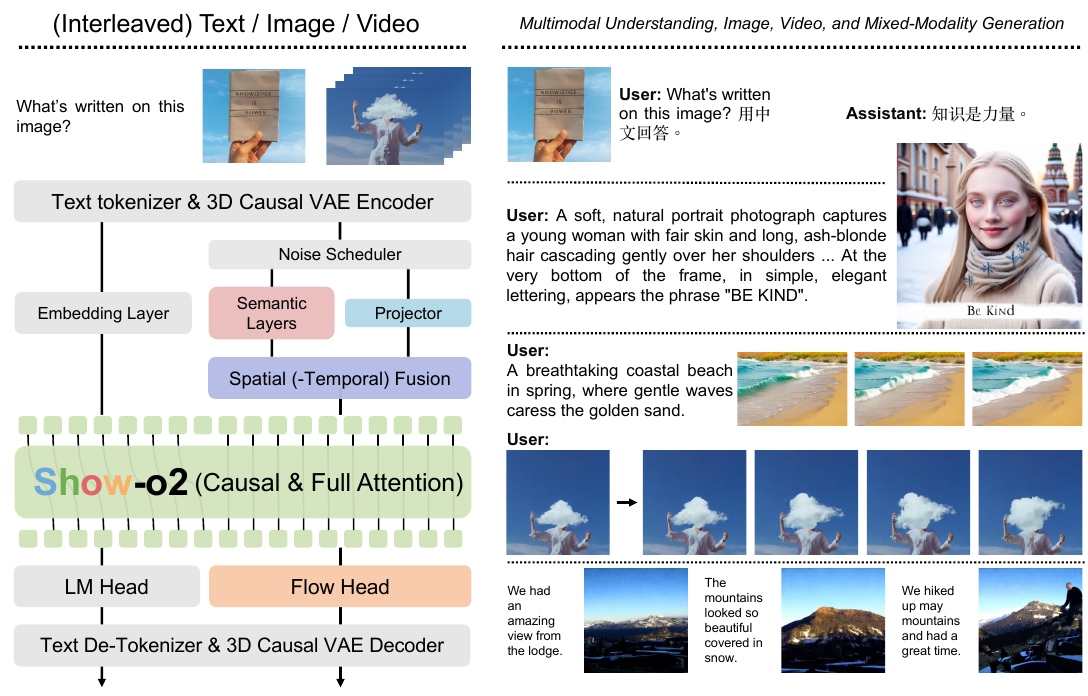}
    \caption{Our approach begins by encoding input texts, images, and videos into continuous embeddings and visual latents. The visual latents are processed through a dual-path extraction and spatial (-temporal) fusion mechanism to construct unified visual representations that are scalable for both multimodal understanding and generation, image and video modalities. These text embeddings and unified visual representations are then structured into a sequence for the base language model, equipped with dedicated heads. Specifically, text tokens are modeled autoregressively by a language head, while image and video latents are handled by a flow head using flow matching. We employ the omni-attention mechanism~\cite{showo,zhou2025transfusion} to enable causal attention along the sequence while maintaining full attention within the unified visual representations. This design empowers our model to effectively tackle tasks such as image/video understanding, generation, and mixed-modality generation.}
    \label{fig:overview}
\end{figure}

\textbf{Overall Architecture.} An overview of our proposed unified model is depicted in Fig.~\ref{fig:overview}. Given (interleaved) texts, images, or videos, a text tokenizer with an embedding layer and a 3D causal VAE encoder accordingly process them into continuous text embeddings and visual latent representations. Subsequently, the visual latent representations undergo a dual-path extraction of spatial (-temporal) fusion to create the unified visual representations. These representations are then structured into a sequence, which is fed into a language model equipped with language and flow heads to model the sequence via autoregressive modeling and flow matching accordingly. Finally, a text de-tokenizer in conjunction with a 3D causal VAE decoder is employed to decode the final output. Next, we will delve into the fundamental design principles behind the unified visual representation and flow head.

\textbf{Unified Visual Representation.} To scalably support image and video modalities, we employ a 3D causal VAE encoder to extract image/video latents. As multimodal understanding and generation differ in feature dependency, we propose a dual-path architecture comprising semantic layers $\mathcal{S}(\cdot)$ to extract high-level representations of rich semantic contextual information and a projector $\mathcal{P}(\cdot)$ to retain complete low-level information from the extracted visual latents. Specifically, semantic layers $\mathcal{S}(\cdot)$ share the same vision transformer blocks of SigLIP~\cite{siglip} with a new $2\times 2$ patch embedding layer. Given $n$ visual latents $\textbf{x}_t=\{x_i\}_{i=1}^n$ at a noise level:
\begin{equation}
\label{eq:add_noise}
    \textbf{x}_t = t\cdot \textbf{x}_{1} + (1-t)\cdot \textbf{x}_0,
\end{equation}
where $\textbf{x}_0 \sim \mathcal{N}(0,1)$ and $t \sim [0,1]$,
we load the pre-trained weights of SigLIP and pre-distill $\mathcal{S}(\cdot)$ as follows:
\begin{equation}
    \mathcal{L}_{\text{distill}} = - \frac{1}{n} \sum \log \text{sim}(\mathcal{S}(\textbf{x}_t), \text{\tt{SigLIP}}(\textbf{X})),
\end{equation}
where $\textbf{X}$ is the input image, $\textbf{\text{\tt{SigLIP}}}(\cdot)$ extracts the image patch features, and $\text{sim}(\cdot)$ indicates the cosine similarity calculator. In this way, semantic layers $\mathcal{S}(\cdot)$ can mimic extracting semantic features from both clean and noised visual latents $\textbf{x}_t$. The projector $\mathcal{P}(\cdot)$ is simply composed of a 2D patch embedding layer. The extracted high- and low-level representations are spatially (and temporally when it comes to videos) fused by concatenating through the feature dimension and applying RMSNorm~\cite{rmsnorm} with two MLP layers to get the unified visual representations $\textbf{u}$:
\begin{equation}
    \textbf{u} = \text{\tt{STF}}(\mathcal{S}(\textbf{x}_t), \mathcal{P}(\textbf{x}_t)),
\end{equation}
where $\text{\tt{STF}}$ indicates the spatial (-temporal) fusion mechanism.
In addition, we prepend a time step $t$ embedding to the unified visual representations for generative modeling. $t$ is set as 1.0 to get time step embedding for the clean image.

We structure the text embeddings and unified visual representations into a sequence following a general interleaved image-text format below:
\begin{equation*}
\label{eq:sequence_format}
    {\fontsize{9}{14}\selectfont \text{[BOS] \{Text\} [BOI / BOV] \{Image / Video\} [EOI / EOV] \{Text\} $\cdots$ [EOS]}}.
\end{equation*}
The sequence format above is flexible and can be adapted to various input types. We adopt the omni-attention mechanism~\cite{showo,zhou2025transfusion} to let the sequence modeling be causal but with full attention within the unified visual representations.

\textbf{Flow Head.} Apart from the language head for text token prediction, we employ a flow head to predict the defined velocity $\textbf{v}_t=\frac{d\textbf{x}_t}{dt}$ via flow matching~\cite{flow, flow2}. Specifically, the flow head simply consists of several transformer layers with time step modulation via the adaLN-Zero blocks, as seen in DiT~\cite{DiT}.

During training, we natively apply next token prediction $\mathcal{L}_\text{NTP}$ to the language head and flow matching $\mathcal{L}_\text{FM}$ to the flow head for predicting velocity, respectively:
\begin{equation}
\label{eq:total_loss}
    \mathcal{L} =  \alpha \mathcal{L}_\text{NTP} + \mathcal{L}_\text{FM}. 
\end{equation} 
\subsection{Training Recipe}
\label{sec:training_strategy}
\begin{wraptable}{r}{0.7\textwidth}
\centering
\vspace{-12pt}
\caption{Trainable components and datasets in the training stages.}
\vspace{-6pt}
\label{tab:training_stages}
\setlength\tabcolsep{6pt}
\resizebox{\linewidth}{!}{ 
\begin{tabular}{llccccccccc}
        \toprule

            & \multirow{2}{*}{Trainable Components} & \multicolumn{3}{c}{Datasets} \\ 
            \cmidrule(lr){3-5}
            & & \# Image-Text & \# Video-Text & \# Interleaved Data \\ 
    	  \midrule
            \multirow{3}{*}{\textbf{Stage-1}} & Projector  & \multirow{3}{*}{66M} & \multirow{3}{*}{\makecell{WebVid~\cite{webvid}\\Pandas~\cite{chen2024panda70m}}} & \multirow{3}{*}{OmniCorpus~\cite{li2024omnicorpus}} \\
            & Spatial (-Temporal) Fusion  &  &  \\   
            &  Flow Head &  & \\
            \midrule
            \multirow{3}{*}{\textbf{Stage-2}} & \multirow{3}{*}{Full Model (w/o VAE)} & 9M HQ Und. & {OpenVid-1M~\cite{nan2024openvid} Gen.} &  {VIST~\cite{visualstorytelling}} \\
            & & 16M HQ Gen. & {1.5M Internal Data Gen.}  & CoMM~\cite{chen2024comm} \\
            & & & 1.6M Video Und. &\\
            
        \bottomrule

\end{tabular}
}
\vspace{-10pt}
\end{wraptable} 
Existing UMMs, such as Show-o~\cite{showo}, Janus-Pro~\cite{janus}, Transfusion~\cite{zhou2025transfusion}, Chameleon~\cite{team2024chameleon}, and Emu3~\cite{wang2024emu3}, are typically trained from LLMs, LMMs, or from scratch. These approaches aim to cultivate visual generative modeling capabilities while preserving language modeling proficiency. However, this process often relies on web-scale, high-quality text corpora, which are prohibitively expensive to collect. Consequently, the lack of such resources can lead to a degradation in language knowledge and modeling performance. To address this challenge, we adopt a two-stage training recipe (as shown in Table~\ref{tab:training_stages}) that effectively retains language knowledge while simultaneously developing visual generation capabilities, without requiring a massive text corpus.

\textbf{Stage-1.} Before the two-stage training, we have pre-distilled the semantic layers $\mathcal{S}(\cdot)$ (implementation details can be found in Section~\ref{sec:experiments}). The first stage only involves trainable components of the projector, spatial (-temporal) fusion, and flow head. In this stage, we train these components using autoregressive modeling and flow matching using around 66M image-text pairs and progressively add interleaved data and video-text pairs.

\textbf{Stage-2.} Subsequently, we tune the full model using 9M high-quality multimodal understanding instruction data, 16M high-quality visual generation data filtered from the 66M image-text pairs, and 1.6M video understanding data.

\textbf{Scaling Up.} After the training of the small-sized model with approximately 1.5B LLM parameters, we resume the pre-trained flow head for the larger model with 7B LLM parameters and introduce a lightweight MLP transformation to align the hidden size, allowing it to quickly adapt to the larger model and converge. 

\section{Experiments}
\label{sec:experiments}
\subsection{Experimental Setup}
\textbf{Datasets.} The curated approximately 66M image-text pairs consist of images with a resolution of at least 512 pixels in width and height. The images are filtered from CC12M~\cite{cc12m}, COYO~\cite{coyo-700m}, LAION-Aesthetic-12M\footnote{https://huggingface.co/datasets/dclure/laion-aesthetics-12m-umap} and AI synthetic data. The images are recaptioned by LMMs except for the synthetic data. The 9M high-quality multimodal understanding instruction data is curated from Densefusion-1M~\cite{li2024DenseFusion}, and LLaVA-OneVision~\cite{llavaonevision}.

\textbf{Implementation Details.} The semantic layers $\mathcal{S}(\cdot)$ are pre-distilled from SigLIP-so400m-patch14-384\footnote{https://huggingface.co/google/siglip-so400m-patch14-384} over 200K iterations, using a batch size of 512 and a cosine-scheduled learning rate of 2e-5. During distillation, Eq.~\ref{eq:add_noise} is applied to the visual latents with only a probability of 0.3 in the last 20K iterations. The input image resolution of 3D causal VAE encoder with $2\times 2$ patch embedding layer is set as $432\times 432$ to get $729=27\times 27$ visual latents, which matches the ones extracted by SigLIP. Once distilled, the semantic layers $\mathcal{S}(\cdot)$ are capable of extracting rich semantic features from both clean and noised visual latents. In statistics, the extracted features from clean visual latents by $\mathcal{S}(\cdot)$ have converged to an average cosine similarity of around 0.9 with those extracted by the original SigLIP on the curated 66M image-text pairs. We interpolate the position embeddings in the bicubic mode when involving other image/video resolutions.

Our models build upon two LLM variants, \emph{i.e.,}  Qwen2.5-1.5B-Instruct~\cite{qwen2.5} and Qwen2.5-7B-Instruct~\cite{qwen2.5}, respectively. We adopt 3D causal VAE proposed in Wan2.1~\cite{wan2025} with 8$\times$ and 4$\times$ spatial and temporal compression, respectively. In stage 1, we first train the 1.5B variant for 150K iterations using AdamW optimizer with a constant learning rate of 0.0001 on the curated 66M image-text pairs in a resolution of $432\times 432$. The context length of single image-text pairs is set as 1024. The total batch sizes for multimodal understanding and generation are 128 and 384, respectively. $\alpha$ in Eq.~\ref{eq:total_loss} is set as 0.2. For visual generation data, the caption is dropped with a probability of 0.1 to enable the classifier-free guidance. This training process roughly takes one and a half days using 64 H100 GPUs. Subsequently, we replace the generation data with 16M high-quality data (filtered from 66M image-text pairs) and continue to train for 40K iterations. In stage 2, we follow the training strategies in LLaVA-OneVision\cite{llavaonevision} to train the 1.5B model using around 9M multimodal instructional and 16M high-quality generation data for a total of around 35K iterations. $\alpha$ in Eq.~\ref{eq:total_loss} is set as 1.0. The stage 2 training process takes around 15 hours. For models with mixed-modality and video generation capabilities, we progressively add video-text and interleaved data in stage 1. For video data, we randomly sample a 2s 480p or 432$\times$432 clips with 17 frames from each video with an interval of 3 frames. The context length at this time is set as 7006. In stage 2, high-quality video-text and interleaved data are added to further improve video and mixed-modality generation capabilities.

To futher improve the image generation and text rendering quality, we further train the small-scale model on images with higher resoluton ($512\times 512$ and $1024\times 1024$) and involve an additional text-rich image data, \emph{i.e.,} a subset of TextAtlas~\cite{textatlas5m}.

Building on the pre-trained image-level Show-o2 models, we enhance their video understanding capabilities by further fine-tuning on 1.6M video samples from \cite{llava-next-video}, together with 1.1M image-level samples from the earlier stage. We adopt the same video training and inference settings as LLaVA-OneVision. The evaluation results are shown in Table~\ref{tab:mmu_video_comparison}.

\begin{table}[t] 
\centering
\caption{Evaluation on multimodal understanding benchmarks. \# Params. indicates the number of parameters of base LLM. * indicates the method uses two distinct models or sets of parameters for multimodal understanding and generation, respectively. $^\dagger$ indicates the Show-o2 models fine-tuned using video understanding data. Und. indicates ``understanding''. Results in \g{gray} indicate the performance of und. only models or models with total parameters more than 13B.}
\label{tab:mmu_comparison}
\resizebox{\linewidth}{!}{ 
\begin{tabular}{llccccccccc}
        \toprule

            \multirow{2}{*}{Types} &  \multirow{2}{*}{Models} &  \multirow{2}{*}{\# Params.} & MME $\uparrow$ & GQA$ \uparrow$ & SEED $\uparrow$ & MMB$\uparrow$ & MMMU $\uparrow$ & MMStar $\uparrow$ & AI2D $\uparrow$\\
            & & & (p) & & (all) & (en) & (val) & \\
    	\midrule 
            \multirow{3}{*}{Und. Only} & \g{LLaVA-v1.5~\cite{llava1.5}} & \g{7B} & \g{1510.7} & \g{62.0} & \g{58.6} & \g{64.3} & \g{-} & \g{-} & \g{-}\\
            & \g{Qwen-VL-Chat~\cite{Qwen-VL}} & \g{7B} & \g{1487.6} & \g{57.5} & \g{58.2} & \g{60.6} & \g{-} & \g{-} & \g{57.7} \\
            & \g{LLaVA-OV~\cite{llavaonevision}} & \g{7B} & \g{1580.0} & \g{-} & \g{-} & \g{80.8} & \g{48.8} & \g{57.5} & \g{81.4} \\
        
        \midrule
            Unify via& \g{NExT-GPT~\cite{showo}} & \g{13B} & \g{-} & \g{-} & \g{57.5} & \g{58.0} & \g{-} & \g{-} & \g{-}  \\
             Assembling & \g{SEED-X~\cite{seed-x}} & \g{17B} & \g{1457.0} & \g{49.1} & \g{66.5} & \g{70.1} & \g{35.6} & \g{-} & \g{-}  \\
            Tailored & MetaMorph~\cite{tong2024metamorph} & 8B &- & - & 71.8 & 75.2 & - & - & -  \\
            Models & \g{TokenFlow-XL$^{*}$~\cite{qu2024tokenflow}} & \g{14B} & \g{1551.1} & \g{62.5} & \g{72.6} & \g{76.8} & \g{43.2} & \g{-} & \g{75.9} \\
            & ILLUME~\cite{ILLUME} & 7B & 1445.3 & - & 72.9 & 75.1 & 38.2 & - & 71.4 \\
        \midrule
            
            \multirow{13}{*}{Native Unified}  & \g{BAGEL~\cite{bagel}} & \g{14B} & \g{1687.0} & \g{-} & \g{-} & \g{85.0} & \g{55.3} & \g{-} & \g{-} \\
            & Show-o~\cite{showo} & 1.3B & 1097.2 & 58.0 & 51.5 & - & 27.4 & - & - \\
            & JanusFlow~\cite{ma2024janusflow} & 1.5B & 1333.1 & 60.3 & 70.5 & 74.9 & 29.3 & - & - \\
            & SynerGen-VL~\cite{synergen-vl} & 2.4B & 1381.0 & - & - & 53.7 & 34.2 & - & - \\
            & Janus-Pro~\cite{janus} & 1.5B & 1444.0 & 59.3 & 68.3 & 75.5 & 36.3 & - & - \\
             \rowcolor{light_green}
            & \textbf{Show-o2 (Ours)} & 1.5B & 1450.9  & 60.0  & 65.6  &  67.4  &  37.1  &  43.4 & 69.0  \\  
            & Emu3~\cite{wang2024emu3} & 8B & - & 60.3 & 68.2 & 58.5 & 31.6 & - & 70.0 \\
            & VILA-U~\cite{vila-u} & 7B & 1401.8 & 60.8 & 59.0 & - & - & - & - \\
            & MUSE-VL~\cite{MUSE-VL} & 7B & - & - & 69.1 & 72.1 & 39.7 & 49.6 & 69.8 \\
            & Liquid~\cite{liquid} & 8B & 1448.0 & 61.1 & - & - & - & - & - \\
            & Janus-Pro~\cite{janus} & 7B & 1567.1 & 62.0 & 72.1 & 79.2 & 41.0 & - & - \\
            & Mogao~\cite{mogao} & 7B & 1592.0 & 60.9 & \textbf{74.6} & 75.0 & 44.2 & 
 - & - \\
             \rowcolor{light_green}
            & \textbf{Show-o2 (Ours)} & 7B & \textbf{1620.5} & \textbf{63.1} & 69.8 &  \textbf{79.3} &  \textbf{48.9} & \textbf{56.6} & \textbf{78.6} \\
        \bottomrule

\end{tabular}
}
\vspace{-4mm}
\end{table}

In the training of our model based on the 7B LLM variant, we resume the flow head pre-trained based on the 1.5B model and additionally train the newly initialized spatial (-temporal) fusion, projector, and MLP transformations for 3K iterations with 2K warm-up steps to align the hidden size and then further train spatial (-temporal) fusion, the projector, MLP transformations, and the flow head together. Following that, we conduct the training stages 1 and 2 in the same manner as those of the 1.5B model. The whole training process of our 7B model takes approximately 2 and a half days using 128 H100 GPUs. We do not include interleaved and video data in the training stages of the larger model due to the huge computational cost and training duration.

\subsection{Multimodal Understanding on Images and Videos}
\vspace{-2mm}
\textbf{Quantitative Results.} Table~\ref{tab:mmu_comparison} highlights the performance of our models on multimodal understanding benchmarks, evaluated across metrics such as MME~\cite{mme}, GQA~\cite{gqa}, SEED-Bench~\cite{seedbench}, MM-Bench~\cite{mmbench}, MMU~\cite{mmmu}, MMStar~\cite{mmstar}, and AI2D~\cite{ai2d}. As shown in the table, both the 1.5B and 7B variants of our model consistently outperform state-of-the-art models across many metrics. For models with similar parameter sizes (1.5B), our model achieves the best scores on MME-p and MMU-val benchmarks while delivering competitive performance on GQA and SEED-Bench metrics. When compared to larger models with approximately 7B parameters, our models surpass state-of-the-art models such as Janus-Pro and even the significantly larger TokenFlow-XL model (14B parameters) in metrics including MME-p, GQA, MMMU-val, MMStar, and AI2D, while maintaining competitive performance on SEED-Bench and MM-Bench. These results underscore the robust perception capabilities of our unified visual representations, demonstrating their effectiveness in multimodal understanding tasks and the promising potential in this domain. In addition, we present the video understanding performance of Show-o2$^\dagger$ in Table~\ref{tab:mmu_video_comparison}.

\textbf{Qualitative Results.} Fig.~\ref{fig:visualization} showcases the multimodal understanding capabilities of our model. As demonstrated, the model excels at answering general-purpose questions about an image. Specifically, it can provide detailed descriptions of an image, count objects, and recognize text within the image. Besides, the model can leverage its world knowledge to offer step-by-step instructions for preparing daily drinks like an avocado milkshake and supports bilingual question-answering, highlighting its versatility and practical utility. Further, our model supports multimodal understanding in both English and Chinese, enabling bilingual capabilities.

\begin{table}[t!]
\setlength\tabcolsep{10pt}
\centering
\caption{Evaluation on video understanding benchmarks. \# Params. denotes the number of parameters in the base LLM, while \# Frames represents the maximum number of video frames used during training and inference. Und. stands for understanding. $^\dagger$ marks the Show-o2 models that have been fine-tuned on video understanding data. All results are reported in terms of zero-shot accuracy.}
\resizebox{\linewidth}{!}{ 
\begin{tabular}{lcccccccc}
    \toprule
    \multirow{2}{*}{\textbf{Model}} & \multirow{2}{*}{\textbf{\# Params.}} & \multirow{2}{*}{\# Frames} & \rotatebox{90}{\textbf{\scriptsize{ActNet-QA}}} &  \rotatebox{90}{\textbf{\scriptsize{MVBench}}} & \rotatebox{90}{\textbf{\scriptsize{NExT-QA}}} & \rotatebox{90}{\textbf{\scriptsize{PerceptionTest}}}  & \rotatebox{90}{\textbf{\scriptsize{LongVideoBench}}} &\rotatebox{90} 
    {\textbf{\scriptsize{VideoMME}}} \\ \cmidrule(l){4-9} 
    & & & test  & test & mc & val   & val & wo/w-subs \\ \midrule
    \textit{Proprietary Und. Only Models} \\
    \rowcolor{mygray}
    GPT-4V~\cite{openai2023gpt4v} & - & - & 57.0 &  43.5 & - & -   & 61.3 & 59.9/63.3 \\
    \rowcolor{mygray}
    GPT-4o~\cite{openai2024gpt4o} &  - &-  & -  & - & - & -    & 66.7 & 71.9/77.2 \\
    \rowcolor{mygray}
    Gemini-1.5-Flash~\cite{team2023gemini} & - & - & 55.3 & -  & - & -   & 61.6  & 70.3/75.0 \\
    \rowcolor{mygray}
    Gemini-1.5-Pro~\cite{team2023gemini} & - & -  & 57.5 &  -  & - & -  & 64.0 & 75.0/81.3 \\ \midrule 
    \textit{Open-source Und. Only Models} \\
    VILA~\cite{lin2024vila} & 40B & -  & 58.0 &  - & 67.9 & 54.0  & -  & 60.1/61.1 \\
    PLLaVA~\cite{xu2024pllava} & 34B & 16 / 16  & 60.9 &  58.1  & - & - & 53.2 & - \\    
    LongVA~\cite{zhang2024long} & 7B & -  & 50.0 &  - & 68.3 & -  & -  & 52.6/54.3 \\
    IXC-2.5~\cite{zhang2024internlm} & 7B & 64 / 64 & 52.8 &  69.1 & 71.0 & 34.4    & - & 55.8/58.8 \\   
    LLaVA-OV~\cite{llavaonevision} & 7B & 32 / 32  & 56.6 & 56.7 & 79.4 & 57.1    & 56.5 & 58.2/61.5 \\
    VideoLLaMA2~\cite{videollama2} & 7B & 16 / 16 &  50.2   & 54.6 & -  & 51.4    & - & 47.9/50.3 \\
    \midrule
    \textit{Unified Multimodal Models} \\
    \rowcolor{light_green}
    \textbf{Show-o2$^\dagger$(Ours)} & 1.5B & 32 / 32 & 52.7 &49.8 & 72.1 & 56.1 & 49.2 & 48.0/51.6 \\
    \rowcolor{light_green}
    \textbf{Show-o2$^\dagger$(Ours)} & 7B & 16 / 32 & 56.4 & 55.8 & 79.0 & 61.9 &55.5 & 57.4/60.9 \\
    \bottomrule
    \end{tabular}}%
\label{tab:mmu_video_comparison}
\vspace{-8mm}
\end{table}

\begin{table}[t]
\centering
\caption{Evaluation on the GenEval~\cite{geneval} benchmark. Gen. denotes ``generation''. \# Params. indicates the number of parameters of base LLM. \# Data. indicates the number of image-text pairs used for visual generation during training. * means the method uses two distinct models for multimodal understanding and generation, respectively. Obj.: Object. Attri.: Attribute. Our results are obtained using rewritten prompts. + indicates the additional data required by the pretrained diffusion models.}
\label{tab:geneval_benchmark}
\resizebox{\linewidth}{!}{  
\begin{tabular}{llcccccccc|c}
        \toprule
    	Type & Method & \# Params. & \# Data & Single Obj. & Two Obj. & Counting  & Colors &  Position & Color Attri. & \textbf{Overall}$\uparrow$ \\
    	\midrule
            \multirow{1}{*}{Gen. Only} & SD3-Medium~\cite{sd3} & - & - & 0.99 & 0.94 & 0.72 & 0.89 & 0.33 & 0.60 & 0.74 \\
            \midrule
            Unifying via & SEED-X~\cite{seed-x} & 17B & 158M+ & 0.97 & 0.58 & 0.26 & 0.80 & 0.19 & 0.14 & 0.49 \\
             Assembling& TokenFlow-XL$^{*}$~\cite{qu2024tokenflow} & 14B & 60M & 0.95 & 0.60 & 0.41 & 0.81 & 0.16 & 0.24 & 0.55 \\
             Tailored& ILLUME~\cite{ILLUME} & 7B & 15M+ & 0.99 & 0.86 & 0.45 & 0.71 & 0.39 & 0.28 & 0.61 \\
            Models& MetaQuery-XL~\cite{metaqueries} & 7B & 28M+ & - & - & - & - & - & - & 0.80 \\
            
            \midrule
             \multirow{10}{*}{Native Unified} & Show-o~\cite{showo} & 1.3B & 2.0B & 0.98 & 0.80 & 0.66 & 0.84 & 0.31 & 0.50 & 0.68 \\
            & Emu3~\cite{wang2024emu3} & 8B & - & - & - & - & - & - & - & 0.66 \\
            & MUSE-VL~\cite{MUSE-VL} & 7B & 24M & & & & & & & 0.57 \\
            & Transfusion~\cite{zhou2025transfusion} & 7B & 3.5B& - & - & - & - & - & - & 0.63 \\
            & D-DiT~\cite{dualdiff} & 2B & 40M & 0.97 & 0.80 & 0.54 & 0.76 & 0.32 & 0.50 & 0.65 \\
            & Janus-Pro~\cite{janus} & 7B & 144M & 0.99 & 0.89 & 0.59 & 0.90 & 0.79 & 0.66 & 0.80 \\
            & BAGEL~\cite{bagel} & 14B & 1600M & 0.98 & 0.95 & 0.84 & 0.95 & 0.78 & 0.77 & 0.88 \\
            & Mogao~\cite{mogao} & 7B & - & 1.00 &  0.97 & 0.83 & 0.93 & 0.84 & 0.80 & \textbf{0.89} \\
             \rowcolor{light_green}
             & \textbf{Show-o2 (Ours)} & 1.5B & 66M & 0.99 & 0.86 & 0.55 & 0.86 & 0.46 & 0.63 & 0.73 \\
             \rowcolor{light_green}
             & \textbf{Show-o2 (Ours)} & 7B & 66M & 1.00  &  0.87 & 0.58  & 0.92  & 0.52  & 0.62 & 0.76  \\
        \bottomrule

\end{tabular}
}
\vspace{-6mm}
\end{table}

\begin{table}[t]
\centering
\caption{Evaluation on the DPG-Bench~\cite{dpgbench} benchmark. Gen. denotes ``generation''. \# Params. indicates the number of parameters of base LLM. \# Data. indicates the number of image-text pairs used for visual generation during training.}
\label{tab:dpg_benchmark}
\setlength\tabcolsep{10pt}
\resizebox{\linewidth}{!}{ 
\begin{tabular}{llccccccc|c}
        \toprule
    	Type & Method & \# Params. & \# Data & Global & Entity & Attribute  & Relation  & Other & \textbf{Overall}$\uparrow$ \\
        \midrule
        \multirow{5}{*}{Gen. Only} & Hunyuan-DiT~\cite{li2024hunyuandit} & 1.5B & - & 84.59 & 80.59 & 88.01 & 74.36 & 86.41 & 78.87 \\
        & Playground v2.5~\cite{Li2024PlaygroundVT} & - & - & 83.06 & 82.59 & 81.20 & 84.08 & 83.50 & 75.47 \\
        & PixArt-$\Sigma$~\cite{chen2024pixartsigma} & - & - & 86.89 & 82.89 & 88.94 & 86.59 & 87.68 & 80.54 \\
        & DALL-E 3~\cite{dalle3} & - & - & 90.97 & 89.61 & 88.39 & 90.58 & 89.83 & 83.50 \\
        & SD3-Medium~\cite{sd3} & 2B & - & 87.90 & 91.01 & 88.83 & 80.70 & 88.68 & 84.08 \\
    	\midrule
        \multirow{5}{*}{Native Unified} & Emu3-DPO~\cite{wang2024emu3} & 8B & - &  - & -  &   &  - &  - & 81.60 \\
        & Janus-Pro~\cite{janus} & 7B & 144M & 86.90 & 88.90 & 89.40 & 89.32 & 89.48 & 84.19 \\
        & Mogao~\cite{mogao} & 7B & - & 82.37 & 90.03 & 88.26 & 93.18 &85.40 & 84.33\\
        \rowcolor{light_green}
        & \textbf{Show-o2 (Ours)} & 1.5B & 66M & 87.53 & 90.38 & 91.34 & 90.30 & 91.21 & 85.02 \\
        \rowcolor{light_green}
        & \textbf{Show-o2 (Ours)} & 7B & 66M & 89.00 & 91.78 & 89.96 & 91.81 & 91.64 & \textbf{86.14} \\
        \bottomrule

\end{tabular}
}
\vspace{-6mm}
\end{table}

\begin{table}[H]
\centering
\caption{Overall quantitative comparison of different methods on OneIG-Bench. Gen. denotes ``generation''. \# Params. indicates the number of parameters of base LLM. \# Data. indicates the number of image-text pairs used for visual generation during training.}
\label{tab:oneig_benchmark}
\setlength\tabcolsep{10pt}
\resizebox{\linewidth}{!}{ 
\begin{tabular}{llccccccc}
        \toprule
    	Type & Method & \# Params. & \# Data & Alignment$\uparrow$ & Text$\uparrow$ & Reasoning$\uparrow$  & Style$\uparrow$ & Diversity$\uparrow$ \\
        \midrule
        \multirow{5}{*}{Gen. Only} & SD3.5-Large\cite{sd3} & 8B & - & 0.809 & 0.629 & 0.294 & 0.353 & 0.225 \\ 
        & Flux.1-dev~\cite{flux2024} & 12B & - & 0.786 & 0.523 & 0.253 & 0.368 & 0.238 \\ 
        & SANA-1.5  (PAG)~\cite{xie2025sana} & 4.8B & - & 0.765 & 0.069 & 0.217 & {0.401} & 0.216 \\ 
        & Lumina-Image 2.0~\cite{lumina2} & 2.6B & 110M &0.819 & 0.106 & 0.270 & 0.354 & 0.216 \\
        &  HiDream-I1-Full~\cite{2025hidreami1} & 17B & - & {0.829} & {0.707} & {0.317} & 0.347 & 0.186 \\
    	\midrule
        \multirow{8}{*}{Unified Models} & Show-o-512~\cite{showo} & 1.3B & 2B & 0.702  & 0.002  &   0.213     & 0.361 & 0.241 \\
        & Janus-Pro~\cite{chen2025janus} & 7B & 144M & 0.553  & 0.001  &   0.139     & 0.276 & 0.365 \\
        & BLIP3-o~\cite{blip3} & 8B & 55M & 0.711  & 0.013  &   0.223      & 0.361 & 0.229 \\
        & BAGEL~\cite{bagel} & 14B & 1600M & 0.769  & 0.244  &   0.173    & 0.367 & 0.251
        \\
        & OmniGen2~\cite{wu2025omnigen2} & 7B & 150M & 0.804 & 0.680 & 0.271 & 0.377 & 0.242 \\ \rowcolor{light_green}
        & \textbf{Show-o2 (Ours)} & 1.5B & 66M & 0.798 & 0.002 & 0.219& 0.317 & 0.186 \\
        \rowcolor{light_green}
        & \textbf{Show-o2-1024$\times$1024 (Ours)} & 1.5B & 66M  & 0.798 & 0.125 & 0.274 & 0.351 & 0.186 \\
        \rowcolor{light_green}
        & \textbf{Show-o2 (Ours)} & 7B & 66M & 0.817 & 0.002 & 0.226 & 0.317 & 0.177\\
        \bottomrule
\end{tabular}
}
\vspace{-6mm}
\end{table}

\vspace{-2mm}
\subsection{Visual Generation}
\begin{figure}[H]
    \centering
    \includegraphics[width=1\linewidth]{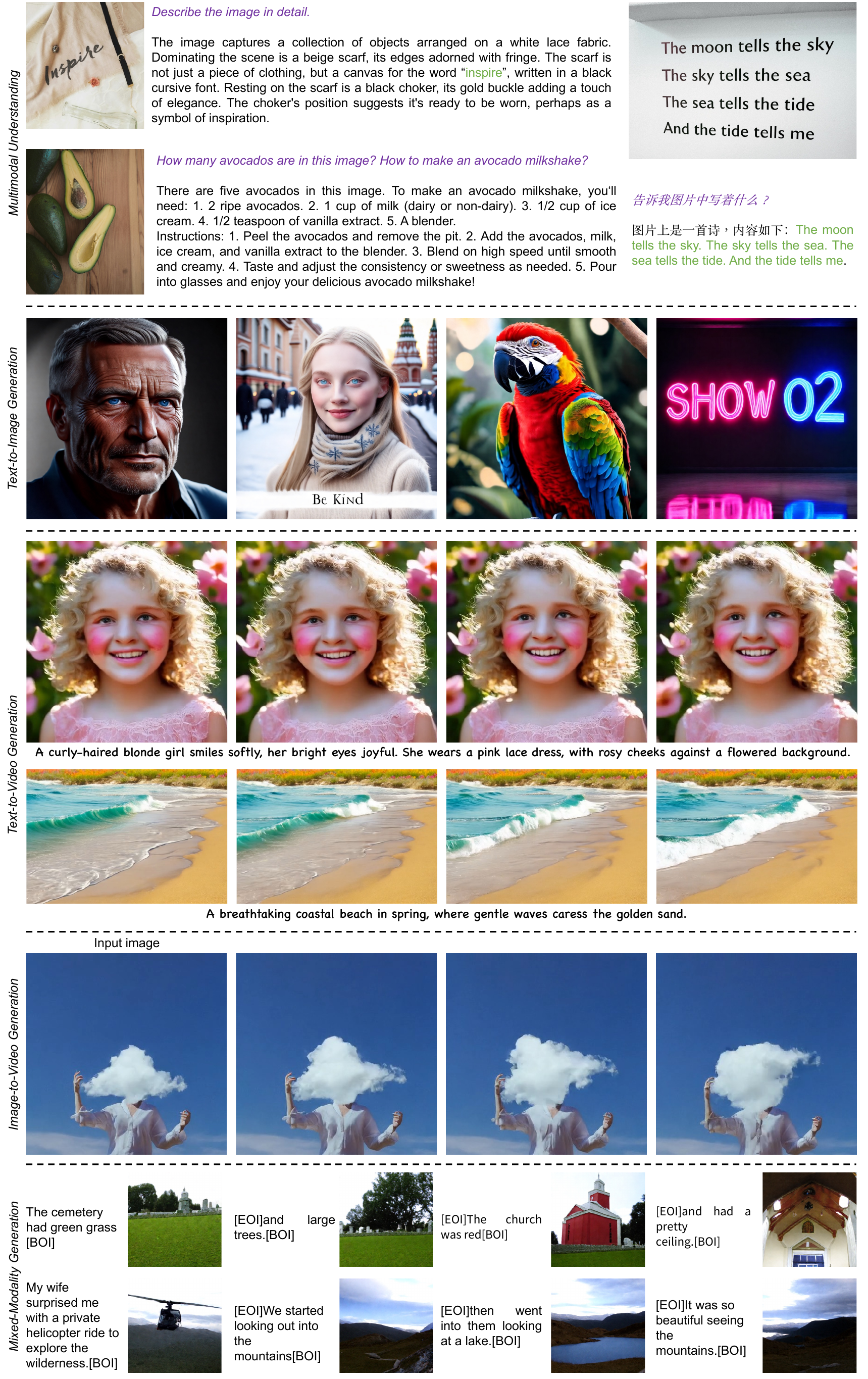}
    \vspace{-20pt}
    \caption{Multimodal understanding and generation examples.}
    \label{fig:visualization}
\end{figure}

\vspace{-2mm}
\textbf{Image Generation.} We compare our model with the state-of-the-art approaches on GenEval~\cite{geneval}, DPG-Bench~\cite{dpgbench}, and OneIG\cite{oneig} benchmarks in Tables~\ref{tab:geneval_benchmark}, \ref{tab:dpg_benchmark}, and \ref{tab:oneig_benchmark}. One can observe that our model surpasses most approaches, including TokenFlow-XL, Show-o, Emu3, and Transfusion, on the GenEval benchmark. Compared to Janus-Pro, which was trained on a significantly larger dataset of 144M image-text pairs, our model achieves promising results with only 66M image-text pairs. On DPG-Bench evaluation, our model has demonstrated the best overall score compared to generation-only models such as SD3-Medium and unified models, including Emu3-DPO and Janus-Pro. On OneIG-Bench, our models also achieve competitive performance. We also show qualitative results in Fig.~\ref{fig:visualization} to illustrate that our model can generate high-quality and realistic images.

\textbf{Video Generation.} We compare our model with the text-to-video and image-to-video generation models in Tables~\ref{tab:t2v_benchmark} and \ref{tab:i2v_benchmark}. One can observe that with only 2B parameters, our model outperforms models such as Show-1, Emu3, and VILA-U with more than 6B parameters. Besides, our model has demonstrated competitive performance compared to CogVideoX and Step-Video-T2V. We also provide qualitative results of the text-to-video and image-to-video generation capability of our model
 in the middle of Fig.~\ref{fig:visualization}. One can observe that, given text prompts or an input image, our model can generate consistent video frames with reasonable motions, such as the smiling girl, lapping waves, and floating clouds.

\subsection{Mixed-Modality Generation}
We demonstrate mixed-modality generation capabilities of our model using downstream task visual storytelling dataset~\cite{visualstorytelling} in Fig.~\ref{fig:visualization}. During fine-tuning, given an interleaved image-text sequence, we apply noise to all images in the sequence with a probability of 0.3. Otherwise, we randomly retain a number of the earlier images in the sequence and only apply noise to the later ones. Benefiting from the general interleaved sequence format mentioned in \ref{eq:sequence_format}, our model can predict the \text{[BOI]} once it begins to generate an image. Upon detecting the \text{[BOI]} token, noises will be appended to the sequence to gradually generate an image. The generated text tokens and images will be served as context to continue generating the following output. Fig.~\ref{fig:visualization} includes two examples demonstrating our model's ability to interleavely generate coherent text and images, vividly narrating a story.

\begin{table}[t]
    \centering
    \setlength{\tabcolsep}{0.095cm}
    \caption{Comparison with text-to-video models on the VBench~\cite{huang2023vbench} benchmark. \# Params. indicates the number of total parameters for video generation including the base LLM and flow head. QS: Quality Score, SS: Semantic Score, SC: Subject Consistency, BC: Background Consistency, TF: Temporal Flickering, MS: Motion Smoothness, DD: Dynamic Degree, AQ: Aesthetic Quality, IQ: Imaging Quality, OC: Object Class, MO: Multiple Objects, HA: Human Action, C: Color, SR: Spatial Relationship, S: Scene, AS: Appearance style, TS: Temporal Style, OC': Overall Consistency.} 
\resizebox{\linewidth}{!}{
\begin{tabular}{lcccccccccccccccccccc}
\toprule
          
          Models &  \# Params.  & Total & QS & SS & SC & BC & TF & MS & DD & AQ & IQ & OC & MO & HA & C & SR & S & AS & TS & OC' \\

\midrule
                 ModelScope~\cite{wang2023modelscope} & 1.7B & 75.75 & 78.05 & 66.54 & 89.87 & 95.29 & 98.28 & 95.79 & 66.39 & 52.06 & 58.57 & 82.25 & 38.98 & 92.40 & 81.72 & 33.68 & 39.26 & 23.39 & 25.37 & 25.67 \\
                LaVie~\cite{wang2023lavie} & 3B & 77.08 & 78.78 & 70.31 & 91.41 & 97.47 & 98.30 & 96.38 & 49.72 & 54.94 & 61.90 & 91.82 & 33.32 & 96.80 & 86.39 & 34.09 & 52.69 & 23.56 & 25.93 & 26.41 \\
                OpenSoraPlan V1.3~\cite{lin2024opensoraplan} & - & 77.23 & 80.14 & 65.62 & 97.79 & 97.24 & 99.20 & 99.05 & 30.28 & 60.42 & 56.21 & 85.56 & 43.58 & 86.80 & 79.30 & 51.61 & 36.73 & 20.03 & 22.47 & 24.47 \\
                Show-1~\cite{show1} & 6B & 78.93 & 80.42 & 72.98 & 95.53 & 98.02 & 99.12 & 98.24 & 44.44 & 57.35 & 58.66 & 93.07 & 45.47 & 95.60 & 86.35 & 53.50 & 47.03 & 23.06 & 25.28 & 27.46 \\
                AnimateDiff-V2~\cite{guo2023animatediff} & - & 80.27 & 82.90 & 69.75 & 95.30 & 97.68 & 98.75 & 97.76 & 40.83 & 67.16 & 70.10 & 90.90 & 36.88 & 92.60 & 87.47 & 34.60 & 50.19 & 22.42 & 26.03 & 27.04 \\
                Gen-2~\cite{Gen2} & - & 80.58 & 82.47 & 73.03 & 97.61 & 97.61 & 99.56 & 99.58 & 18.89 & 66.96 & 67.42 & 90.92 & 55.47 & 89.20 & 89.49 & 66.91 & 48.91 & 19.34 & 24.12 & 26.17 \\
                Pika-1.0~\cite{pika1} & - & 80.69 & 82.92 & 71.77 & 96.94 & 97.36 & 99.74 & 99.50 & 47.50 & 62.04 & 61.87 & 88.72 & 43.08 & 86.20 & 90.57 & 61.03 & 49.83 & 22.26 & 24.22 & 25.94 \\
                VideoCrafter-2.0~\cite{chen2024videocrafter2} & - & 80.44 & 82.20 & 73.42 & 96.85 & 98.22 & 98.41 & 97.73 & 42.50 & 63.13 & 67.22 & 92.55 & 40.66 & 95.00 & 92.92 & 35.86 & 55.29 & 25.13 & 25.84 & 28.23 \\
                CogVideoX~\cite{yang2024cogvideox} & 5B & 81.61 & 82.75 & 77.04 & 96.23 & 96.52 & 98.66 & 96.92 & 70.97 & 61.98 & 62.90 & 85.23 & 62.11 & 99.40 & 82.81 & 66.35 & 53.20 & 24.91 & 25.38 & 27.59 \\
                Kling~\cite{kling} & - & 81.85 & 83.39 & 75.68 & 98.33 & 97.60 & 99.30 & 99.40 & 46.94 & 61.21 & 65.62 & 87.24 & 68.05 & 93.40 & 89.90 & 73.03 & 50.86 & 19.62 & 24.17 & 26.42 \\
                Step-Video-T2V~\cite{stept2v} & 30B & 81.83 & 84.46 & 71.28 & 98.05 & 97.67 & 99.40 & 99.08 & 53.06 & 61.23 & 70.63 & 80.56 & 50.55 & 94.00 & 88.25 & 71.47 & 24.38 & 23.17 & 26.01 & 27.12 \\
                Gen-3~\cite{Gen3} & -& 82.32 & 84.11 & 75.17 & 97.10 & 96.62 & 98.61 & 99.23 & 60.14 & 63.34 & 66.82 & 87.81 & 53.64 & 96.40 & 80.90 & 65.09 & 54.57 & 24.31 & 24.71 & 26.69 \\
                \midrule
                        Emu3~\cite{wang2024emu3} & 8B & 80.96 & - & - & 95.32 & 97.69 & - & 98.93 &79.27 & 59.64 & - & 86.17 & 44.64 & 77.71 & - & 68.73 & 37.11 & 20.92 & - & - \\
                        VILA-U~\cite{vila-u} & 7B & 74.01 & 76.26 & 65.04 & - &- &- &- &- &- &- &- &- &- &- &- &- & - &- &- \\
                        HaploOmni~\cite{xiao2025haploomni} & 9B & 78.10 & - & - & 96.40 & 97.60 &- & 96.80 & 65.30 &- &- &- &- &- &- &- & 34.60 & - &- &- \\
                        \rowcolor{light_green}
                        \textbf{Show-o2 (Ours)} & 2B & 81.34 & 82.10 & 78.31 & 97.28 & 96.78 & 97.68 & 98.25 & 40.83 & 65.15 & 67.06 & 94.81 & 76.01 & 95.20 & 80.89 & 62.61 & 57.67& 23.29 & 25.27& 27.00 \\
\bottomrule
\end{tabular}
}
\vspace{-6pt}

    \label{tab:t2v_benchmark}
\end{table}

\begin{table}[t]
    \centering
    \setlength{\tabcolsep}{0.095cm}
    \caption{Comparison with image-to-video models on  the VBench~\cite{huang2023vbench} benchmark.} 
\resizebox{\linewidth}{!}{
\begin{tabular}{lccccccccccccc}
\toprule
          Models &  \makecell{I2V\\ Subject} & \makecell{I2V\\Background} & \makecell{Camera\\Motion} & \makecell{Subject\\Consistency} & \makecell{Background\\Consistency} & \makecell{Temporal\\Flickering} & \makecell{Motion\\Smoothness} & \makecell{Dynamic\\Degree} &  \makecell{Aesthetic \\ Quality} & \makecell{Imaging\\Quality} \\

\midrule
DynamiCrafter-1024~\cite{xing2023dynamicrafter} & 96.71& 96.05& 35.44& 95.69& 97.38& 97.63& 97.38& 47.40& 66.46& 69.34 \\
SEINE-512x320~\cite{chen2023seine} & 94.85& 94.02& 23.36& 94.20& 97.26& 96.72& 96.68& 34.31& 58.42& 70.97 \\
I2VGen-XL~\cite{zhang2023i2vgen} & 96.74& 95.44& 13.32& 96.36& 97.93& 98.48& 98.31& 24.96& 65.33& 69.85 \\
Animate-Anything~\cite{animate-anything} & 98.54& 96.88& 12.56& 98.90& 98.19& 98.14& 98.61& 2.68& 67.12& 72.09 \\
ConsistI2V~\cite{ren2024consisti2v} & 94.69& 94.57& 33.60& 95.27& 98.28& 97.56& 97.38& 18.62& 59.00& 66.92 \\
VideoCrafter-I2V~\cite{chen2023videocrafter1} & 90.97& 90.51& 33.58& 97.86& 98.79& 98.19& 98.00& 22.60& 60.78& 71.68 \\
SVD-XT-1.1~\cite{svd} & 97.51& 97.62& - & 95.42& 96.77& 99.17& 98.12& 43.17& 60.23& 70.23 \\
MarDini~\cite{mardini} & 98.78 & 96.46 & - & -& -& -& - & - & - & - \\
\midrule
\rowcolor{light_green}
\textbf{Show-o2 (Ours)} & 96.94 & 98.83 & 28.41 & 93.83 & 97.45 & - & 97.76 & 25.85 & 61.92 & 69.87 \\
\bottomrule
\end{tabular}
}
\vspace{-6pt}

    \label{tab:i2v_benchmark}
\end{table}

\subsection{Ablation Studies}
\begin{wraptable}{r}{.4\linewidth}
    \vspace{-12pt}
	\caption{\small{Impact of spatial (-temporal) fusion.}}
    \resizebox{\linewidth}{!}{ 
	\begin{tabular}{lcccccccc}
        \toprule
        & MME$-\text{p}$ $\uparrow$ & GQA $\uparrow$ & POPE $\uparrow$ &FID-5K $\downarrow$ \\ 
      \midrule
        w/o Fusion & 1164.7 & 56.2 & 82.6 & 21.8 \\
        w Fusion &  \textbf{1187.8} & \textbf{57.6} & 82.6 & \textbf{20.5} \\
        \bottomrule
    \end{tabular}}
    \label{tab:stf_ablation}
    \vspace{-6pt}
\end{wraptable}
We show the pilot study results in Table~\ref{tab:stf_ablation}, which validated the effect of spatial (-temporal) fusion on multimodal understanding and generation performance. For efficiency, we adopt LLaMA-3.2-1B as the base language model and use only around 1M multimodal understanding data and the ImageNet-1K generation data~\cite{imagenet}. Under the same training settings, there are improvements in terms of both multimodal understanding and generation metrics, including MME-p, GQA, and FID-5K. This validates that the involved semantic and low-level features in the fusion mechanism would potentially help both the multimodal generation and understanding capabilities to some extent.

\begin{wraptable}{r}{.5\linewidth}
	\caption{\small{Effect of CFG guidance and inference steps.}}
    \resizebox{\linewidth}{!}{ 
	\begin{tabular}{cc|ccccccc}
        \toprule

       CFG guidance & Inference steps & GenEval & DPG-Bench \\ 
      \midrule
        2.5 & 50 & 0.65  & 81.6    \\
        5.0 &  50 & 0.71 & 83.9 \\
        7.5 &  50 & 0.71 & 84.8   \\
        10 & 50 & 0.71 & \textbf{85.0} \\
        \midrule
        7.5 &  25 & 0.71 & 84.6 \\
        7.5 & 100 & \textbf{0.73} & 84.7 \\
        \bottomrule

    \end{tabular}}
    \label{tab:cfg_step_ablation}
    \vspace{-6pt}
\end{wraptable}
We perform ablation studies to examine the effect of classifier-free guidance (CFG) and inference steps on the performance using the 1.5B model. As shown in Table~\ref{tab:cfg_step_ablation}, increasing the CFG guidance scale and inference steps (in a range) would potentially improve the GenEval and DPG-Bench scores. However, the improvements of the GenEval score are not significant when the CFG guidance is set as larger than 5.0. 

\begin{wraptable}{r}{.35\linewidth}
    \vspace{-12pt}
	\caption{\small{Effect of training stages.}}
    \vspace{-6pt}
    \resizebox{\linewidth}{!}{ 
	\begin{tabular}{cc|ccccccc}
        \toprule

       Stage-1 & Stage-2 & GenEval & DPG-Bench \\ 
      \midrule
        \checkmark &   &  0.63  &  83.28    \\
        \checkmark & \checkmark &  \textbf{0.73}  &  \textbf{84.70}    \\
        \bottomrule

    \end{tabular}}
    \label{tab:training_stage_ablation}
    \vspace{-6pt}
\end{wraptable}
Table~\ref{tab:training_stage_ablation} provides the effect of training stages on the generation performance on the GenEval and DPG-Bench benchmarks. One can observe that stage-2 training consistently and significantly improves both metrics, which validates the importance of the second stage.

\begin{table}[h]
    \vspace{-12pt}
	\caption{Impact of training recipe on text-only performance. One-stage training denotes full-parameter-co-training on image-text pairs and the text-only RefinedWeb~\cite{refinedweb} data. Note that the curated multimodal understanding data consists of text-only instructional data. We perform the evaluation under the same setting using the \href{https://github.com/EleutherAI/lm-evaluation-harness}{\text{lm-evaluation-harness tool}}.}
    \resizebox{\linewidth}{!}{ 
	\begin{tabular}{lcccccccc}
        \toprule
        Models & \# Params. & Training Recipe & MMLU  & GPQA & GSM8K  & HumanEval \\ 
      \midrule
        Qwen2.5 Instruct~\cite{qwen2.5} & 1.5B & - & 60.20 $\pm$ 0.39 & 28.12 $\pm$ 2.13 & 51.86 $\pm$ 1.38 & 35.37 $\pm$ 3.74 \\
    Show-o2 (Ours)& 1.5B & One-stage training with RefinedWeb & 28.25 $\pm$ 0.38 & 25.00 $\pm$ 2.05 & 4.55 $\pm$ 0.57 & 3.05 $\pm$ 1.35 \\
    \rowcolor{light_green}
    Show-o2 (Ours)& 1.5B & Our two-stage training & 56.75 $\pm$ 1.37 & 29.24 $\pm$ 2.15 & 49.43 $\pm$ 1.38 & 35.54 $\pm$ 3.70 \\
    \midrule
    Qwen2.5 Instruct~\cite{qwen2.5} & 7B & - & 71.75 $\pm$ 0.36 & 32.37 $\pm$ 2.21 & 82.49 $\pm$ 1.05 & 65.24 $\pm$ 3.73 \\
    Show-o2 (Ours)& 7B& One-stage training with RefinedWeb & 28.43 $\pm$ 0.21 & 26.34 $\pm$ 2.08 & 1.52 $\pm$ 0.34 & 4.01 $\pm$ 1.25 \\
    \rowcolor{light_green}
    Show-o2 (Ours)& 7B & Our two-stage training & 70.73 $\pm$ 0.36 & 31.47 $\pm$ 2.22 & 75.28 $\pm$ 1.19 & 70.73 $\pm$ 3.56 \\
        \bottomrule
    \end{tabular}}
    \label{tab:two_stage_ablation}
    \vspace{-6pt}
\end{table}
Table~\ref{tab:two_stage_ablation} shows that our models effectively preserve language knowledge and achieve performance comparable to the original Qwen2.5-1.5B and Qwen2.5-7B Instruct models. In contrast, direct one-stage full-parameter-co-training with textual data such as RefinedWeb results in substantial performance degradation, highlighting the necessity of the two-stage training approach when high-quality corpora are unavailable.

\begin{wraptable}{r}{0.7\linewidth}
    \vspace{-12pt}
	\caption{\small{Impact of image token count on chart, text, and document VQA.}}
    \vspace{-4pt}
    \resizebox{\linewidth}{!}{ 
	\begin{tabular}{lcccccccc}
        \toprule
        Models & \# Params. & \# Image tokens & ChartQA & DocVQA$_\text{val}$ & InfoVQA$_\text{val}$ & TextVQA$_\text{val}$ \\
        \midrule
        LLaVA-OV & 7B & 729 & 56.24 & 62.71 & 39.59 & 66.19 \\
        \midrule
        \rowcolor{light_green}
        \textbf{Show-o2} & 7B & 729 & 48.00 & 59.34 & 42.31 & 62.92 \\
        \rowcolor{light_green}
        \textbf{Show-o2} & 7B & 5$\times$729  & 66.92 & 77.26 & 45.80 & 71.54 \\
        \bottomrule
    \end{tabular}}
    \label{tab:textvqa_ablation}
    \vspace{-6pt}
\end{wraptable}
As shown in Table~\ref{tab:textvqa_ablation}, our ablation study reveals that increasing the number of image tokens significantly boosts performance across all tasks, even though the model was trained with a fixed image resolution. Using the AnyRes strategy at inference time consistently improves results, highlighting the benefit of higher token counts for capturing fine-grained details. When compared to the baseline LLaVA-OV-7B, our model achieves comparable results on DocVQA, InfoVQA, and TextVQA validation sets, but underperforms on ChartQA. We attribute this gap to the limited chart-related data available during semantic layer distillation, which constrains the model’s ability to capture chart-specific information. We believe that incorporating more OCR and document-centric data into the distillation process will further strengthen the unified model’s OCR and document understanding capabilities.

\section{Limitations and Broader Impacts}
\label{sec:limitations_and_broader_impacts}
We found that our model is not good at rendering text on the image. We investigated our generation datasets and observed that the proportion of images with rendered texts is relatively small, which potentially leads to bad text rendering. In addition, the generated images will lack details of the small objects because of the limited image resolution. To address this limitation, as outlined in the implementation details, we have enhanced the model by training it on higher resolution data and incorporating image datasets rich in textual information.

Our models possess the ability to generate text and images, which may carry the risk of unintended misuse, such as creating fake information or profiles. Additionally, our large-scale dataset includes content featuring celebrities and copyrighted materials, which could potentially result in intellectual property infringement.

\section{Conclusion}
This paper proposed native unified multimodal models, \emph{i.e.,} Show-o2, scalable for multimodal understanding and generation, image and video modalities, by integrating 3D causal VAE, autoregressive modeling, and flow matching. A dual-path of spatial (-temporal) fusion mechanism guided the construction of unified visual representations with both high- and low-level features. A two-stage training recipe enables effective learning of unified capabilties, resulting in a versatile model capable of handling diverse tasks, including multimodal understanding and image/video generation. Extensive experiments demonstrate the model's state-of-the-art performance across various benchmarks. 

\begin{ack}
We thank Haozhe Liu for his valuable input and discussions throughout this project. We are also grateful to Meng Wei and Weihao Wang for their assistance in preparing and organizing the datasets for image and video generation.
\end{ack}








\clearpage
\appendix

\section{Technical Appendices and Supplementary Material}
\begin{table}[t]
\centering
\caption{Comparative analysis of selected unified multimodal models based on the utilization of visual representations and type of unified modeling for multimodal understanding and generation. In this context, \textbf{native und. \& gen.} refers to the direct decoding of output sequences into texts, images, and videos, as opposed to serving as conditions for decoding using external pre-trained decoders like Stable Diffusion. $^{*}$ indicates the method uses two distinct models for multimodal understanding and generation, respectively.}
\label{tab:model_compar_complete}
\resizebox{\linewidth}{!}{ 
\begin{tabular}{lcccccccc}
        \toprule
        
            \multirow{2}{*}{Methods} & \multicolumn{3}{c}{Und. \& Gen. Representation} & \multicolumn{3}{c}{Type of Unified Modeling} \\
            \cmidrule(lr){2-4}  \cmidrule(lr){5-7}
            & Unified & Decoupled & Support Video & Native Und. \& Gen. & Assembling Tailored Models & Paradigm \\
    	  \midrule
             Chameleon~\cite{team2024chameleon} & \checkmark & & \usym{2613} & \checkmark &  & AR \\
            Show-o~\cite{showo} & \checkmark & & \usym{2613} &\checkmark & & AR + Diff.\\
            Transfusion~\cite{zhou2025transfusion} & \checkmark & & \usym{2613} &\checkmark & & AR + Diff.\\
            VILA-U~\cite{vila-u} & \checkmark & & \checkmark & \checkmark & & AR \\
            Emu3~\cite{wang2024emu3} & \checkmark & & \checkmark &\checkmark & & AR \\
            MonoFormer~\cite{monoformer} & \checkmark & & \usym{2613} &\checkmark & & AR + Diff. \\
            Dual-Diffusion~\cite{dualdiff} & \checkmark & & \usym{2613} &\checkmark & & Diff. \\
            SynerGen-VL~\cite{synergen-vl} & \checkmark & & \usym{2613} &\checkmark & & AR \\
            MMAR~\cite{MMAR} & \checkmark & & \usym{2613} &\checkmark & & AR + MAR\\
            MUSE-VL~\cite{MUSE-VL} & \checkmark & & \usym{2613} &\checkmark & & AR\\
            Orthus~\cite{kou2024orthus} & \checkmark &  &\usym{2613} & \checkmark & & AR + Diff. \\
            Liquid~\cite{liquid} & \checkmark & & \usym{2613} &\checkmark & & AR \\
            LlamaFusion~\cite{lmfusion} & \checkmark & & \usym{2613}  & \checkmark &  & AR + Diff.\\
            UGen~\cite{ugen} & \checkmark & & \usym{2613} &\checkmark &  & AR \\
            UniDisc~\cite{swerdlow2025unidisc} & \checkmark & & \usym{2613} &\checkmark & & Diff. 
            \\
            UniToken~\cite{unitoken} & \checkmark & & \usym{2613} &\checkmark & & AR \\
            Harmon~\cite{wu2025harmon} & \checkmark & & \usym{2613} &\checkmark &  & AR+MAR \\
            DualToken~\cite{dualtoken} & \checkmark & & \usym{2613} &\checkmark &  & AR \\
            UniTok~\cite{unitok} & \checkmark & & \usym{2613} &\checkmark & & AR \\
            
            Selftok~\cite{selftok} & \checkmark & & \usym{2613} &\checkmark & & AR 
            \\
            Muddit~\cite{shi2025muddit} & \checkmark & & \usym{2613} &\checkmark & & Diff. 
            \\
            MMaDA~\cite{yang2025mmada} & \checkmark & & \usym{2613} &\checkmark & & Diff. 
            \\
             HaploOmni~\cite{xiao2025haploomni} & \checkmark & & \checkmark &\checkmark & & AR + Diff.
            \\
            TokLIP~\cite{lin2025toklip} & \checkmark & & \usym{2613} &\checkmark & & AR \\
            \rowcolor{light_green}
             Show-o2 (Ours) & \checkmark & & \checkmark & \checkmark & & AR + Diff. \\
            \midrule
            Janus-Series~\cite{janus,ma2024janusflow,chen2025janus} & & \checkmark &\usym{2613} & \checkmark & & AR (+Diff.)\\
            VARGPT~\cite{vargpt} &  & \checkmark & \usym{2613} &\checkmark & & AR 
            \\UnidFluid~\cite{unifluid} & & \checkmark &\usym{2613} & \checkmark & & AR + MAR \\ 
            OmniMamba~\cite{OmniMamba} & & \checkmark &\usym{2613} & \checkmark & & AR \\ 
            
            Mogao~\cite{mogao} & & \checkmark & \usym{2613} & \checkmark & & AR + Diff. \\ 
            BAGEL~\cite{bagel} & & \checkmark & \checkmark & \checkmark & & AR + Diff. \\ 
            Fudoki~\cite{wang2025fudokidiscreteflowbasedunified} &   & \checkmark & \usym{2613}& \checkmark & & Diff. \\
            UniGen~\cite{tian2025unigen} &   & \checkmark & \usym{2613}& \checkmark & & AR + Diff. \\
            \midrule
            NExT-GPT~\cite{wu2023next} & & \checkmark& \checkmark & & \checkmark & AR + Diff.\\
            CoDI~\cite{CoDI} & & \checkmark& \checkmark & & \checkmark & AR + Diff.\\
            DreamLLM~\cite{dreamllm} & & \checkmark& \usym{2613} & & \checkmark & AR + Diff.\\
            SEED-X~\cite{seed-x} & & \checkmark& \usym{2613} & & \checkmark & AR + Diff.\\
            MIO~\cite{mio} & & \checkmark& \checkmark & & \checkmark & AR + Diff.\\
            CoDI-2~\cite{codi2} & & \checkmark& \checkmark & & \checkmark & AR + Diff.\\
            MetaMorph~\cite{tong2024metamorph} &  & \checkmark& \usym{2613} & & \checkmark & AR + Diff. \\
            ILLUME~\cite{ILLUME} &  & \checkmark & \usym{2613}& & \checkmark & AR + Diff. \\
            ILLUME+~\cite{illume_plus} &  & \checkmark& \usym{2613} & & \checkmark & AR + Diff. \\
            MetaQueries~\cite{metaqueries} &  & \checkmark & \usym{2613}& & \checkmark & AR + Diff. \\
            Nexus-Gen~\cite{Nexus-Gen} &  & \checkmark & \usym{2613}& & \checkmark & AR + Diff.\\
            Ming-Lite-Uni~\cite{Ming-Lite-Uni} & & \checkmark & \usym{2613}& & \checkmark & AR + Diff.\\
            BLIP3-o~\cite{blip3} &  & \checkmark & \usym{2613}& & \checkmark & AR + Diff. \\
            OpenUni~\cite{wu2025openuni} &  & \checkmark & \usym{2613}& & \checkmark & AR + Diff. \\
            UniWorld~\cite{lin2025uniworld} &  & \checkmark & \usym{2613}& & \checkmark & AR + Diff. \\
            Ming-Omni~\cite{Mingomni2025} &  & \checkmark & \checkmark & & \checkmark & AR + Diff. \\
            Pisces~\cite{xu2025pisces} &  & \checkmark & \usym{2613} & & \checkmark & AR + Diff. \\
            \midrule
            TokenFlow$^{*}$~\cite{qu2024tokenflow} & \checkmark & & \usym{2613} & & \checkmark & AR\\
            SemHiTok$^{*}$~\cite{SemHiTok} & \checkmark & & \usym{2613} & & \checkmark & AR \\
        \bottomrule

\end{tabular}
}
\end{table}

\subsection{More Qualitative Results}
\begin{figure}[H]
    \centering
    \includegraphics[width=1\linewidth]{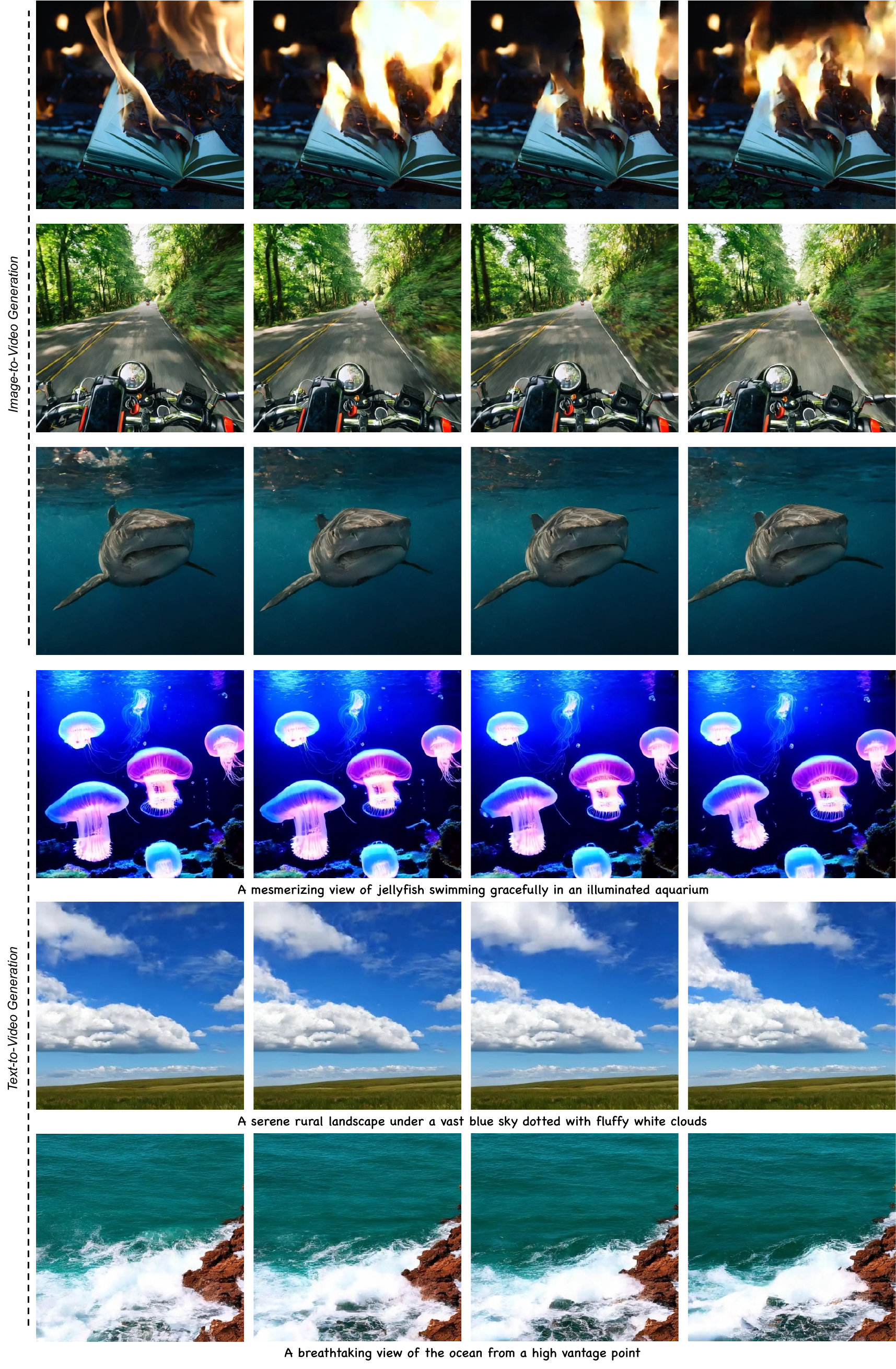}
    \caption{Text-to-video and image-to-video generation examples.}
    \label{fig:visualization2}
\end{figure}

\subsection{Text Prompts}
We provide the text prompts for image generation used in Fig.~\ref{fig:visualization} below:

``Hyper-detailed image of a mature man with short, graying hair and deep blue eyes. He has a rugged, weathered face with a strong jawline and a slight beard. His expression is thoughtful and introspective. The lighting is dramatic, highlighting the contours of his face. The photo is in 8K resolution, capturing every wrinkle and pore.
''

``A soft, natural portrait photograph captures a young woman with fair skin and long, ash-blonde hair cascading gently over her shoulders, her striking light blue eyes subtly enhanced with natural makeup and a gentle, calm smile playing on her lips. She wears a cozy, cream-colored winter sweater and a delicate woolen scarf adorned with subtle snowflake patterns, positioned slightly off-center, creating a sense of relaxed elegance. Behind her, a softly blurred snowy Moscow street scene unfolds, with traditional architecture and the diffused, golden glow of a winter afternoon contributing to a serene and contemplative atmosphere. At the very bottom of the frame, in simple, elegant lettering, appears the phrase "BE KIND".
''

``A vibrant, highly detailed close-up of a colorful parrot perched on a branch, featuring intricate feather textures, vivid colors (red, blue, green, yellow), and a tropical rainforest background. The parrot's eyes are sharp and expressive, with a natural glint of light. The image is photorealistic, ultra HD (8K resolution), with soft natural lighting and a shallow depth of field, creating a blurred bokeh effect in the background. The scene is peaceful and lush, showcasing the beauty of nature.
''

``A dark, moody room with a glowing neon sign on the wall that spells out 'SHOW O2' in bold, vibrant pink and blue colors. The neon light reflects softly on the polished concrete floor, creating a futuristic and artistic vibe.
''

\newpage

    {
        \small
        
        \bibliographystyle{ieee_fullname}
        \bibliography{ref}

\begin{thebibliography}{100}\itemsep=-1pt

\bibitem{Gen2}
Gen-2.
\newblock Accessed September 25, 2023 [Online] \url{https://research.runwayml.com/gen2}, 2023.

\bibitem{pika1}
Pika 1.0.
\newblock Accessed December 28, 2023 [Online] \url{https://www.pika.art/}, 2023.

\bibitem{Gen3}
Gen-3.
\newblock Accessed June 17, 2024 [Online] \url{https://runwayml.com/research/introducing-gen-3-alpha}, 2024.

\bibitem{kling}
Kling.
\newblock Accessed June 6, 2024 [Online] \url{https://klingai.kuaishou.com/}, 2024.

\bibitem{Mingomni2025}
Inclusion AI.
\newblock Ming-omni: A unified multimodal model for perception and generation, 2025.

\bibitem{Qwen-VL}
Jinze Bai, Shuai Bai, Shusheng Yang, Shijie Wang, Sinan Tan, Peng Wang, Junyang Lin, Chang Zhou, and Jingren Zhou.
\newblock Qwen-vl: A versatile vision-language model for understanding, localization, text reading, and beyond.
\newblock {\em arXiv preprint arXiv:2308.12966}, 2023.

\bibitem{Qwen2.5-VL}
Shuai Bai, Keqin Chen, Xuejing Liu, Jialin Wang, Wenbin Ge, Sibo Song, Kai Dang, Peng Wang, Shijie Wang, Jun Tang, Humen Zhong, Yuanzhi Zhu, Mingkun Yang, Zhaohai Li, Jianqiang Wan, Pengfei Wang, Wei Ding, Zheren Fu, Yiheng Xu, Jiabo Ye, Xi Zhang, Tianbao Xie, Zesen Cheng, Hang Zhang, Zhibo Yang, Haiyang Xu, and Junyang Lin.
\newblock Qwen2.5-vl technical report.
\newblock {\em arXiv preprint arXiv:2502.13923}, 2025.

\bibitem{webvid}
Max Bain, Arsha Nagrani, G{\"u}l Varol, and Andrew Zisserman.
\newblock Frozen in time: A joint video and image encoder for end-to-end retrieval.
\newblock In {\em IEEE International Conference on Computer Vision}, 2021.

\bibitem{uvit}
Fan Bao, Shen Nie, Kaiwen Xue, Yue Cao, Chongxuan Li, Hang Su, and Jun Zhu.
\newblock All are worth words: A vit backbone for diffusion models.
\newblock In {\em CVPR}, 2023.

\bibitem{dalle3}
James Betker, Gabriel Goh, Li Jing, † TimBrooks, Jianfeng Wang, Linjie Li, † LongOuyang, † JuntangZhuang, † JoyceLee, † YufeiGuo, † WesamManassra, † PrafullaDhariwal, † CaseyChu, † YunxinJiao, and Aditya Ramesh.
\newblock Improving image generation with better captions.

\bibitem{svd}
Andreas Blattmann, Tim Dockhorn, Sumith Kulal, Daniel Mendelevitch, Maciej Kilian, Dominik Lorenz, Yam Levi, Zion English, Vikram Voleti, Adam Letts, et~al.
\newblock Stable video diffusion: Scaling latent video diffusion models to large datasets.
\newblock {\em arXiv preprint arXiv:2311.15127}, 2023.

\bibitem{coyo-700m}
Minwoo Byeon, Beomhee Park, Haecheon Kim, Sungjun Lee, Woonhyuk Baek, and Saehoon Kim.
\newblock Coyo-700m: Image-text pair dataset.
\newblock \url{https://github.com/kakaobrain/coyo-dataset}, 2022.

\bibitem{oneig}
Jingjing Chang, Yixiao Fang, Peng Xing, Shuhan Wu, Wei Cheng, Rui Wang, Xianfang Zeng, Gang Yu, and Hai-Bao Chen.
\newblock Oneig-bench: Omni-dimensional nuanced evaluation for image generation.
\newblock {\em arXiv preprint arxiv:2506.07977}, 2025.

\bibitem{cc12m}
Soravit Changpinyo, Piyush Sharma, Nan Ding, and Radu Soricut.
\newblock Conceptual 12m: Pushing web-scale image-text pre-training to recognize long-tail visual concepts.
\newblock In {\em CVPR}, pages 3558--3568, 2021.

\bibitem{chen2023videocrafter1}
Haoxin Chen, Menghan Xia, Yingqing He, Yong Zhang, Xiaodong Cun, Shaoshu Yang, Jinbo Xing, Yaofang Liu, Qifeng Chen, Xintao Wang, Chao Weng, and Ying Shan.
\newblock Videocrafter1: Open diffusion models for high-quality video generation.
\newblock {\em arXiv preprint arXiv:2310.19512}, 2023.

\bibitem{chen2024videocrafter2}
Haoxin Chen, Yong Zhang, Xiaodong Cun, Menghan Xia, Xintao Wang, Chao Weng, and Ying Shan.
\newblock Videocrafter2: Overcoming data limitations for high-quality video diffusion models, 2024.

\bibitem{chen2024pixartsigma}
Junsong Chen, Chongjian Ge, Enze Xie, Yue Wu, Lewei Yao, Xiaozhe Ren, Zhongdao Wang, Ping Luo, Huchuan Lu, and Zhenguo Li.
\newblock Pixart-sigma: Weak-to-strong training of diffusion transformer for 4k text-to-image generation, 2024.

\bibitem{blip3}
Jiuhai Chen, Zhiyang Xu, Xichen Pan, Yushi Hu, Can Qin, Tom Goldstein, Lifu Huang, Tianyi Zhou, Saining Xie, Silvio Savarese, et~al.
\newblock Blip3-o: A family of fully open unified multimodal models-architecture, training and dataset.
\newblock {\em arXiv preprint arXiv:2505.09568}, 2025.

\bibitem{pixart}
Junsong Chen, Jincheng Yu, Chongjian Ge, Lewei Yao, Enze Xie, Zhongdao Wang, James~T. Kwok, Ping Luo, Huchuan Lu, and Zhenguo Li.
\newblock Pixart-{\(\alpha\)}: Fast training of diffusion transformer for photorealistic text-to-image synthesis.
\newblock In {\em {ICLR}}. OpenReview.net, 2024.

\bibitem{sharegpt4v}
Lin Chen, Jisong Li, Xiaoyi Dong, Pan Zhang, Conghui He, Jiaqi Wang, Feng Zhao, and Dahua Lin.
\newblock Sharegpt4v: Improving large multi-modal models with better captions.
\newblock {\em arXiv preprint arXiv:2311.12793}, 2023.

\bibitem{mmstar}
Lin Chen, Jinsong Li, Xiaoyi Dong, Pan Zhang, Yuhang Zang, Zehui Chen, Haodong Duan, Jiaqi Wang, Yu Qiao, Dahua Lin, et~al.
\newblock Are we on the right way for evaluating large vision-language models?
\newblock {\em arXiv preprint arXiv:2403.20330}, 2024.

\bibitem{chen2020generative}
Mark Chen, Alec Radford, Rewon Child, Jeffrey Wu, Heewoo Jun, David Luan, and Ilya Sutskever.
\newblock Generative pretraining from pixels.
\newblock In {\em ICML}, pages 1691--1703, 2020.

\bibitem{chen2024panda70m}
Tsai-Shien Chen, Aliaksandr Siarohin, Willi Menapace, Ekaterina Deyneka, Hsiang-wei Chao, Byung~Eun Jeon, Yuwei Fang, Hsin-Ying Lee, Jian Ren, Ming-Hsuan Yang, and Sergey Tulyakov.
\newblock Panda-70m: Captioning 70m videos with multiple cross-modality teachers.
\newblock In {\em Proceedings of the IEEE/CVF Conference on Computer Vision and Pattern Recognition}, 2024.

\bibitem{chen2024comm}
Wei Chen, Lin Li, Yongqi Yang, Bin Wen, Fan Yang, Tingting Gao, Yu Wu, and Long Chen.
\newblock Comm: A coherent interleaved image-text dataset for multimodal understanding and generation.
\newblock {\em arXiv preprint arXiv:2406.10462}, 2024.

\bibitem{chen2023seine}
Xinyuan Chen, Yaohui Wang, Lingjun Zhang, Shaobin Zhuang, Xin Ma, Jiashuo Yu, Yali Wang, Dahua Lin, Yu Qiao, and Ziwei Liu.
\newblock Seine: Short-to-long video diffusion model for generative transition and prediction.
\newblock {\em arXiv preprint arXiv:2310.20700}, 2023.

\bibitem{janus}
Xiaokang Chen, Zhiyu Wu, Xingchao Liu, Zizheng Pan, Wen Liu, Zhenda Xie, Xingkai Yu, and Chong Ruan.
\newblock Janus-pro: Unified multimodal understanding and generation with data and model scaling.
\newblock {\em arXiv preprint arXiv:2501.17811}, 2025.

\bibitem{chen2025janus}
Xiaokang Chen, Zhiyu Wu, Xingchao Liu, Zizheng Pan, Wen Liu, Zhenda Xie, Xingkai Yu, and Chong Ruan.
\newblock Janus-pro: Unified multimodal understanding and generation with data and model scaling.
\newblock {\em arXiv preprint arXiv:2501.17811}, 2025.

\bibitem{SemHiTok}
Zisheng Chen, Chunwei Wang, Xiuwei Chen, Hang Xu, Jianhua Han, and Xiaodan Liang.
\newblock Semhitok: {A} unified image tokenizer via semantic-guided hierarchical codebook for multimodal understanding and generation.
\newblock {\em arXiv preprint arXiv:2503.06764}, 2025.

\bibitem{internvl}
Zhe Chen, Weiyun Wang, Yue Cao, Yangzhou Liu, Zhangwei Gao, Erfei Cui, Jinguo Zhu, Shenglong Ye, Hao Tian, Zhaoyang Liu, et~al.
\newblock Expanding performance boundaries of open-source multimodal models with model, data, and test-time scaling.
\newblock {\em arXiv preprint arXiv:2412.05271}, 2024.

\bibitem{videollama2}
Zesen Cheng, Sicong Leng, Hang Zhang, Yifei Xin, Xin Li, Guanzheng Chen, Yongxin Zhu, Wenqi Zhang, Ziyang Luo, Deli Zhao, and Lidong Bing.
\newblock Videollama 2: Advancing spatial-temporal modeling and audio understanding in video-llms, 2024.

\bibitem{animate-anything}
Zuozhuo Dai, Zhenghao Zhang, Yao Yao, Bingxue Qiu, Siyu Zhu, Long Qin, and Weizhi Wang.
\newblock Fine-grained open domain image animation with motion guidance.
\newblock {\em arXiv preprint arXiv:2311.12886}, 2023.

\bibitem{bagel}
Chaorui Deng, Deyao Zhu, Kunchang Li, Chenhui Gou, Feng Li, Zeyu Wang, Shu Zhong, Weihao Yu, Xiaonan Nie, Ziang Song, et~al.
\newblock Emerging properties in unified multimodal pretraining.
\newblock {\em arXiv preprint arXiv:2505.14683}, 2025.

\bibitem{imagenet}
Jia Deng, Wei Dong, Richard Socher, Li-Jia Li, Kai Li, and Li Fei-Fei.
\newblock Imagenet: A large-scale hierarchical image database.
\newblock In {\em CVPR}, pages 248--255, 2009.

\bibitem{diao2024EVE}
Haiwen Diao, Yufeng Cui, Xiaotong Li, Yueze Wang, Huchuan Lu, and Xinlong Wang.
\newblock Unveiling encoder-free vision-language models.
\newblock {\em arXiv preprint arXiv:2406.11832}, 2024.

\bibitem{diao2025EVEv2}
Haiwen Diao, Xiaotong Li, Yufeng Cui, Yueze Wang, Haoge Deng, Ting Pan, Wenxuan Wang, Huchuan Lu, and Xinlong Wang.
\newblock Evev2: Improved baselines for encoder-free vision-language models.
\newblock {\em arXiv preprint arXiv:2502.06788}, 2025.

\bibitem{dreamllm}
Runpei Dong, Chunrui Han, Yuang Peng, Zekun Qi, Zheng Ge, Jinrong Yang, Liang Zhao, Jianjian Sun, Hongyu Zhou, Haoran Wei, Xiangwen Kong, Xiangyu Zhang, Kaisheng Ma, and Li Yi.
\newblock Dream{LLM}: Synergistic multimodal comprehension and creation.
\newblock In {\em ICLR}, 2024.

\bibitem{sd3}
Patrick Esser, Sumith Kulal, Andreas Blattmann, Rahim Entezari, Jonas M{\"u}ller, Harry Saini, Yam Levi, Dominik Lorenz, Axel Sauer, Frederic Boesel, et~al.
\newblock Scaling rectified flow transformers for high-resolution image synthesis.
\newblock In {\em ICML}, 2024.

\bibitem{unifluid}
Lijie Fan, Luming Tang, Siyang Qin, Tianhong Li, Xuan Yang, Siyuan Qiao, Andreas Steiner, Chen Sun, Yuanzhen Li, Tao Zhu, et~al.
\newblock Unified autoregressive visual generation and understanding with continuous tokens.
\newblock {\em arXiv preprint arXiv:2503.13436}, 2025.

\bibitem{mme}
Chaoyou Fu, Peixian Chen, Yunhang Shen, Yulei Qin, Mengdan Zhang, Xu Lin, Zhenyu Qiu, Wei Lin, Jinrui Yang, Xiawu Zheng, Ke Li, Xing Sun, and Rongrong Ji.
\newblock {MME:} {A} comprehensive evaluation benchmark for multimodal large language models.
\newblock {\em CoRR}, abs/2306.13394, 2023.

\bibitem{seed-x}
Yuying Ge, Sijie Zhao, Jinguo Zhu, Yixiao Ge, Kun Yi, Lin Song, Chen Li, Xiaohan Ding, and Ying Shan.
\newblock Seed-x: Multimodal models with unified multi-granularity comprehension and generation.
\newblock {\em arXiv preprint arXiv:2404.14396}, 2024.

\bibitem{geneval}
Dhruba Ghosh, Hannaneh Hajishirzi, and Ludwig Schmidt.
\newblock Geneval: An object-focused framework for evaluating text-to-image alignment.
\newblock In {\em NeurIPS}, 2023.

\bibitem{Ming-Lite-Uni}
Biao Gong, Cheng Zou, Dandan Zheng, Hu Yu, Jingdong Chen, Jianxin Sun, Junbo Zhao, Jun Zhou, Kaixiang Ji, Lixiang Ru, et~al.
\newblock Ming-lite-uni: Advancements in unified architecture for natural multimodal interaction.
\newblock {\em arXiv preprint arXiv:2505.02471}, 2025.

\bibitem{guo2023animatediff}
Yuwei Guo, Ceyuan Yang, Anyi Rao, Zhengyang Liang, Yaohui Wang, Yu Qiao, Maneesh Agrawala, Dahua Lin, and Bo Dai.
\newblock Animatediff: Animate your personalized text-to-image diffusion models without specific tuning.
\newblock {\em International Conference on Learning Representations}, 2024.

\bibitem{2025hidreami1}
HiDream-ai.
\newblock Hidream-i1.
\newblock \url{https://github.com/HiDream-ai/HiDream-I1}, 2025.

\bibitem{dpgbench}
Xiwei Hu, Rui Wang, Yixiao Fang, Bin Fu, Pei Cheng, and Gang Yu.
\newblock Ella: Equip diffusion models with llm for enhanced semantic alignment, 2024.

\bibitem{illume_plus}
Runhui Huang, Chunwei Wang, Junwei Yang, Guansong Lu, Yunlong Yuan, Jianhua Han, Lu Hou, Wei Zhang, Lanqing Hong, Hengshuang Zhao, and Hang Xu.
\newblock Illume+: Illuminating unified mllm with dual visual tokenization and diffusion refinement.
\newblock {\em arXiv preprint arXiv:2504.01934}, 2025.

\bibitem{visualstorytelling}
Ting-Hao~K. Huang, Francis Ferraro, Nasrin Mostafazadeh, Ishan Misra, Jacob Devlin, Aishwarya Agrawal, Ross Girshick, Xiaodong He, Pushmeet Kohli, Dhruv Batra, et~al.
\newblock Visual storytelling.
\newblock In {\em 15th Annual Conference of the North American Chapter of the Association for Computational Linguistics (NAACL 2016)}, 2016.

\bibitem{huang2023vbench}
Ziqi Huang, Yinan He, Jiashuo Yu, Fan Zhang, Chenyang Si, Yuming Jiang, Yuanhan Zhang, Tianxing Wu, Qingyang Jin, Nattapol Chanpaisit, Yaohui Wang, Xinyuan Chen, Limin Wang, Dahua Lin, Yu Qiao, and Ziwei Liu.
\newblock {VBench}: Comprehensive benchmark suite for video generative models.
\newblock In {\em Proceedings of the IEEE/CVF Conference on Computer Vision and Pattern Recognition}, 2024.

\bibitem{gqa}
Drew~A. Hudson and Christopher~D. Manning.
\newblock {GQA:} {A} new dataset for real-world visual reasoning and compositional question answering.
\newblock In {\em {CVPR}}, pages 6700--6709. Computer Vision Foundation / {IEEE}, 2019.

\bibitem{unitoken}
Yang Jiao, Haibo Qiu, Zequn Jie, Shaoxiang Chen, Jingjing Chen, Lin Ma, and Yu-Gang Jiang.
\newblock Unitoken: Harmonizing multimodal understanding and generation through unified visual encoding.
\newblock {\em arXiv preprint arXiv:2504.04423}, 2025.

\bibitem{ai2d}
Aniruddha Kembhavi, Mike Salvato, Eric Kolve, Minjoon Seo, Hannaneh Hajishirzi, and Ali Farhadi.
\newblock A diagram is worth a dozen images.
\newblock In {\em Computer Vision--ECCV 2016: 14th European Conference, Amsterdam, The Netherlands, October 11--14, 2016, Proceedings, Part IV 14}, pages 235--251. Springer, 2016.

\bibitem{kondratyuk2023videopoet}
Dan Kondratyuk, Lijun Yu, Xiuye Gu, Jos{\'e} Lezama, Jonathan Huang, Rachel Hornung, Hartwig Adam, Hassan Akbari, Yair Alon, Vighnesh Birodkar, et~al.
\newblock Videopoet: A large language model for zero-shot video generation.
\newblock {\em arXiv preprint arXiv:2312.14125}, 2023.

\bibitem{kou2024orthus}
Siqi Kou, Jiachun Jin, Chang Liu, Ye Ma, Jian Jia, Quan Chen, Peng Jiang, and Zhijie Deng.
\newblock Orthus: Autoregressive interleaved image-text generation with modality-specific heads.
\newblock {\em arXiv preprint arXiv:2412.00127}, 2024.

\bibitem{flux2024}
Black~Forest Labs.
\newblock Flux.
\newblock \url{https://github.com/black-forest-labs/flux}, 2024.

\bibitem{seedbench}
Bohao Li, Rui Wang, Guangzhi Wang, Yuying Ge, Yixiao Ge, and Ying Shan.
\newblock Seed-bench: Benchmarking multimodal llms with generative comprehension.
\newblock {\em arXiv preprint arXiv:2307.16125}, 2023.

\bibitem{llavaonevision}
Bo Li, Yuanhan Zhang, Dong Guo, Renrui Zhang, Feng Li, Hao Zhang, Kaichen Zhang, Yanwei Li, Ziwei Liu, and Chunyuan Li.
\newblock Llava-onevision: Easy visual task transfer.
\newblock {\em arXiv preprint arXiv:2408.03326}, 2024.

\bibitem{Li2024PlaygroundVT}
Daiqing Li, Aleks Kamko, Ehsan Akhgari, Ali Sabet, Linmiao Xu, and Suhail Doshi.
\newblock Playground v2.5: Three insights towards enhancing aesthetic quality in text-to-image generation.
\newblock {\em ArXiv}, abs/2402.17245, 2024.

\bibitem{synergen-vl}
Hao Li, Changyao Tian, Jie Shao, Xizhou Zhu, Zhaokai Wang, Jinguo Zhu, Wenhan Dou, Xiaogang Wang, Hongsheng Li, Lewei Lu, and Jifeng Dai.
\newblock Synergen-vl: Towards synergistic image understanding and generation with vision experts and token folding.
\newblock {\em arXiv preprint arXiv:2412.09604}, 2024.

\bibitem{ARPG}
Haopeng Li, Jinyue Yang, Guoqi Li, and Huan Wang.
\newblock Autoregressive image generation with randomized parallel decoding.
\newblock {\em arXiv preprint arXiv:2503.10568}, 2025.

\bibitem{li2024omnicorpus}
Qingyun Li, Zhe Chen, Weiyun Wang, Wenhai Wang, Shenglong Ye, Zhenjiang Jin, et~al.
\newblock Omnicorpus: A unified multimodal corpus of 10 billion-level images interleaved with text.
\newblock In {\em The Thirteenth International Conference on Learning Representations}, 2025.

\bibitem{mar}
Tianhong Li, Yonglong Tian, He Li, Mingyang Deng, and Kaiming He.
\newblock Autoregressive image generation without vector quantization.
\newblock {\em arXiv preprint arXiv:2406.11838}, 2024.

\bibitem{li2024DenseFusion}
Xiaotong Li, Fan Zhang, Haiwen Diao, Yueze Wang, Xinlong Wang, and Ling-Yu Duan.
\newblock Densefusion-1m: Merging vision experts for comprehensive multimodal perception.
\newblock {\em 2407.08303}, 2024.

\bibitem{dualdiff}
Zijie Li, Henry Li, Yichun Shi, Amir~Barati Farimani, Yuval Kluger, Linjie Yang, and Peng Wang.
\newblock Dual diffusion for unified image generation and understanding.
\newblock {\em arXiv preprint arXiv:2501.00289}, 2024.

\bibitem{li2024hunyuandit}
Zhimin Li, Jianwei Zhang, Qin Lin, Jiangfeng Xiong, Yanxin Long, Xinchi Deng, Yingfang Zhang, Xingchao Liu, Minbin Huang, Zedong Xiao, Dayou Chen, Jiajun He, Jiahao Li, Wenyue Li, Chen Zhang, Rongwei Quan, Jianxiang Lu, Jiabin Huang, Xiaoyan Yuan, Xiaoxiao Zheng, Yixuan Li, Jihong Zhang, Chao Zhang, Meng Chen, Jie Liu, Zheng Fang, Weiyan Wang, Jinbao Xue, Yangyu Tao, Jianchen Zhu, Kai Liu, Sihuan Lin, Yifu Sun, Yun Li, Dongdong Wang, Mingtao Chen, Zhichao Hu, Xiao Xiao, Yan Chen, Yuhong Liu, Wei Liu, Di Wang, Yong Yang, Jie Jiang, and Qinglin Lu.
\newblock Hunyuan-dit: A powerful multi-resolution diffusion transformer with fine-grained chinese understanding, 2024.

\bibitem{mogao}
Chao Liao, Liyang Liu, Xun Wang, Zhengxiong Luo, Xinyu Zhang, Wenliang Zhao, Jie Wu, Liang Li, Zhi Tian, and Weilin Huang.
\newblock Mogao: An omni foundation model for interleaved multi-modal generation.
\newblock {\em arXiv preprint arXiv:2505.05472}, 2025.

\bibitem{lin2024opensoraplan}
Bin Lin, Yunyang Ge, Xinhua Cheng, Zongjian Li, Bin Zhu, Shaodong Wang, Xianyi He, Yang Ye, Shenghai Yuan, Liuhan Chen, et~al.
\newblock Open-sora plan: Open-source large video generation model.
\newblock {\em arXiv preprint arXiv:2412.00131}, 2024.

\bibitem{lin2025uniworld}
Bin Lin, Zongjian Li, Xinhua Cheng, Yuwei Niu, Yang Ye, Xianyi He, Shenghai Yuan, Wangbo Yu, Shaodong Wang, Yunyang Ge, et~al.
\newblock Uniworld: High-resolution semantic encoders for unified visual understanding and generation.
\newblock {\em arXiv preprint arXiv:2506.03147}, 2025.

\bibitem{lin2025toklip}
Haokun Lin, Teng Wang, Yixiao Ge, Yuying Ge, Zhichao Lu, Ying Wei, Qingfu Zhang, Zhenan Sun, and Ying Shan.
\newblock Toklip: Marry visual tokens to clip for multimodal comprehension and generation.
\newblock {\em arXiv preprint arXiv:2505.05422}, 2025.

\bibitem{lin2024vila}
Ji Lin, Hongxu Yin, Wei Ping, Pavlo Molchanov, Mohammad Shoeybi, and Song Han.
\newblock Vila: On pre-training for visual language models.
\newblock In {\em Proceedings of the IEEE/CVF Conference on Computer Vision and Pattern Recognition}, pages 26689--26699, 2024.

\bibitem{flow}
Yaron Lipman, Ricky T.~Q. Chen, Heli Ben-Hamu, Maximilian Nickel, and Matthew Le.
\newblock Flow matching for generative modeling.
\newblock In {\em The Eleventh International Conference on Learning Representations}, 2023.

\bibitem{llava1.5}
Haotian Liu, Chunyuan Li, Yuheng Li, and Yong~Jae Lee.
\newblock Improved baselines with visual instruction tuning.
\newblock In {\em CVPR}, pages 26296--26306, 2024.

\bibitem{llava}
Haotian Liu, Chunyuan Li, Qingyang Wu, and Yong~Jae Lee.
\newblock Visual instruction tuning.
\newblock {\em NeurIPS}, 36, 2024.

\bibitem{mardini}
Haozhe Liu, Shikun Liu, Zijian Zhou, Mengmeng Xu, Yanping Xie, Xiao Han, Juan~C. Pérez, Ding Liu, Kumara Kahatapitiya, Menglin Jia, Jui-Chieh Wu, Sen He, Tao Xiang, Jürgen Schmidhuber, and Juan-Manuel Pérez-Rúa.
\newblock Mardini: Masked autoregressive diffusion for video generation at scale.
\newblock {\em arXiv preprint arXiv:2410.20280}, 2024.

\bibitem{flow2}
Xingchao Liu, Chengyue Gong, and Qiang Liu.
\newblock Flow straight and fast: Learning to generate and transfer data with rectified flow.
\newblock {\em arXiv preprint arXiv:2209.03003}, 2022.

\bibitem{mmbench}
Yuan Liu, Haodong Duan, Yuanhan Zhang, Bo Li, Songyang Zhang, Wangbo Zhao, Yike Yuan, Jiaqi Wang, Conghui He, Ziwei Liu, et~al.
\newblock Mmbench: Is your multi-modal model an all-around player?
\newblock In {\em European conference on computer vision}, pages 216--233. Springer, 2024.

\bibitem{uio2}
Jiasen Lu, Christopher Clark, Sangho Lee, Zichen Zhang, Savya Khosla, Ryan Marten, Derek Hoiem, and Aniruddha Kembhavi.
\newblock Unified-io 2: Scaling autoregressive multimodal models with vision, language, audio, and action.
\newblock {\em arXiv preprint arXiv:2312.17172}, 2023.

\bibitem{unitok}
Chuofan Ma, Yi Jiang, Junfeng Wu, Jihan Yang, Xin Yu, Zehuan Yuan, Bingyue Peng, and Xiaojuan Qi.
\newblock Unitok: A unified tokenizer for visual generation and understanding.
\newblock {\em arXiv preprint arXiv:2502.20321}, 2025.

\bibitem{stept2v}
Guoqing Ma, Haoyang Huang, Kun Yan, Liangyu Chen, Nan Duan, Shengming Yin, Changyi Wan, Ranchen Ming, Xiaoniu Song, Xing Chen, Yu Zhou, Deshan Sun, Deyu Zhou, Jian Zhou, Kaijun Tan, Kang An, Mei Chen, Wei Ji, Qiling Wu, Wen Sun, Xin Han, Yanan Wei, Zheng Ge, Aojie Li, Bin Wang, Bizhu Huang, Bo Wang, Brian Li, Changxing Miao, Chen Xu, Chenfei Wu, Chenguang Yu, Dapeng Shi, Dingyuan Hu, Enle Liu, Gang Yu, Ge Yang, Guanzhe Huang, Gulin Yan, Haiyang Feng, Hao Nie, Haonan Jia, Hanpeng Hu, Hanqi Chen, Haolong Yan, Heng Wang, Hongcheng Guo, Huilin Xiong, Huixin Xiong, Jiahao Gong, Jianchang Wu, Jiaoren Wu, Jie Wu, Jie Yang, Jiashuai Liu, Jiashuo Li, Jingyang Zhang, Junjing Guo, Junzhe Lin, Kaixiang Li, Lei Liu, Lei Xia, Liang Zhao, Liguo Tan, Liwen Huang, Liying Shi, Ming Li, Mingliang Li, Muhua Cheng, Na Wang, Qiaohui Chen, Qinglin He, Qiuyan Liang, Quan Sun, Ran Sun, Rui Wang, Shaoliang Pang, Shiliang Yang, Sitong Liu, Siqi Liu, Shuli Gao, Tiancheng Cao, Tianyu Wang, Weipeng Ming, Wenqing He, Xu Zhao, Xuelin Zhang,
  Xianfang Zeng, Xiaojia Liu, Xuan Yang, Yaqi Dai, Yanbo Yu, Yang Li, Yineng Deng, Yingming Wang, Yilei Wang, Yuanwei Lu, Yu Chen, Yu Luo, Yuchu Luo, Yuhe Yin, Yuheng Feng, Yuxiang Yang, Zecheng Tang, Zekai Zhang, Zidong Yang, Binxing Jiao, Jiansheng Chen, Jing Li, Shuchang Zhou, Xiangyu Zhang, Xinhao Zhang, Yibo Zhu, Heung-Yeung Shum, and Daxin Jiang.
\newblock Step-video-t2v technical report: The practice, challenges, and future of video foundation model, 2025.

\bibitem{ma2024janusflow}
Yiyang Ma, Xingchao Liu, Xiaokang Chen, Wen Liu, Chengyue Wu, Zhiyu Wu, Zizheng Pan, Zhenda Xie, Haowei Zhang, Xingkai yu, Liang Zhao, Yisong Wang, Jiaying Liu, and Chong Ruan.
\newblock Janusflow: Harmonizing autoregression and rectified flow for unified multimodal understanding and generation, 2024.

\bibitem{nan2024openvid}
Kepan Nan, Rui Xie, Penghao Zhou, Tiehan Fan, Zhenheng Yang, Zhijie Chen, Xiang Li, Jian Yang, and Ying Tai.
\newblock Openvid-1m: A large-scale high-quality dataset for text-to-video generation.
\newblock {\em arXiv preprint arXiv:2407.02371}, 2024.

\bibitem{openai2023gpt4v}
OpenAI.
\newblock Gpt-4v.
\newblock \url{https://openai.com/index/gpt-4v-system-card/}, 2023.

\bibitem{openai2024gpt4o}
OpenAI.
\newblock Hello gpt-4o.
\newblock \url{https://openai.com/index/hello-gpt-4o/}, 2024.

\bibitem{metaqueries}
Xichen Pan, Satya~Narayan Shukla, Aashu Singh, Zhuokai Zhao, Shlok~Kumar Mishra, Jialiang Wang, Zhiyang Xu, Jiuhai Chen, Kunpeng Li, Felix Juefei-Xu, Ji Hou, and Saining Xie.
\newblock Transfer between modalities with metaqueries.
\newblock {\em arXiv preprint arXiv:2504.06256}, 2025.

\bibitem{randar}
Ziqi Pang, Tianyuan Zhang, Fujun Luan, Yunze Man, Hao Tan, Kai Zhang, William~T. Freeman, and Yu-Xiong Wang.
\newblock Randar: Decoder-only autoregressive visual generation in random orders.
\newblock {\em arXiv preprint arXiv:2412.01827}, 2024.

\bibitem{DiT}
William Peebles and Saining Xie.
\newblock Scalable diffusion models with transformers.
\newblock {\em arXiv preprint arXiv:2212.09748}, 2022.

\bibitem{peebles2023scalable}
William Peebles and Saining Xie.
\newblock Scalable diffusion models with transformers.
\newblock In {\em ICCV}, pages 4195--4205, 2023.

\bibitem{refinedweb}
Guilherme Penedo, Quentin Malartic, Daniel Hesslow, Ruxandra Cojocaru, Hamza Alobeidli, Alessandro Cappelli, Baptiste Pannier, Ebtesam Almazrouei, and Julien Launay.
\newblock The refinedweb dataset for falcon {LLM:} outperforming curated corpora with web data only.
\newblock In {\em NeurIPS}, 2023.

\bibitem{lumina2}
Qi Qin, Le Zhuo, Yi Xin, Ruoyi Du, Zhen Li, Bin Fu, Yiting Lu, Xinyue Li, Dongyang Liu, Xiangyang Zhu, Will Beddow, Erwann Millon, Wenhai~Wang Victor~Perez, Yu Qiao, Bo Zhang, Xiaohong Liu, Hongsheng Li, Chang Xu, and Peng Gao.
\newblock Lumina-image 2.0: A unified and efficient image generative framework, 2025.

\bibitem{qu2024tokenflow}
Liao Qu, Huichao Zhang, Yiheng Liu, Xu Wang, Yi Jiang, Yiming Gao, Hu Ye, Daniel~K Du, Zehuan Yuan, and Xinglong Wu.
\newblock Tokenflow: Unified image tokenizer for multimodal understanding and generation.
\newblock {\em arXiv preprint arXiv:2412.03069}, 2024.

\bibitem{clip}
Alec Radford, Jong~Wook Kim, Chris Hallacy, Aditya Ramesh, Gabriel Goh, Sandhini Agarwal, Girish Sastry, Amanda Askell, Pamela Mishkin, Jack Clark, Gretchen Krueger, and Ilya Sutskever.
\newblock Learning transferable visual models from natural language supervision.
\newblock In {\em {ICML}}, pages 8748--8763, 2021.

\bibitem{ren2024consisti2v}
Weiming Ren, Harry Yang, Ge Zhang, Cong Wei, Xinrun Du, Stephen Huang, and Wenhu Chen.
\newblock Consisti2v: Enhancing visual consistency for image-to-video generation.
\newblock {\em arXiv preprint arXiv:2402.04324}, 2024.

\bibitem{rombach2022high}
Robin Rombach, Andreas Blattmann, Dominik Lorenz, Patrick Esser, and Bj{\"o}rn Ommer.
\newblock High-resolution image synthesis with latent diffusion models.
\newblock In {\em CVPR}, pages 10684--10695, 2022.

\bibitem{seawead2025seaweed}
Team Seawead, Ceyuan Yang, Zhijie Lin, Yang Zhao, Shanchuan Lin, Zhibei Ma, Haoyuan Guo, Hao Chen, Lu Qi, Sen Wang, et~al.
\newblock Seaweed-7b: Cost-effective training of video generation foundation model.
\newblock {\em arXiv preprint arXiv:2504.08685}, 2025.

\bibitem{shi2025muddit}
Qingyu Shi, Jinbin Bai, Zhuoran Zhao, Wenhao Chai, Kaidong Yu, Jianzong Wu, Shuangyong Song, Yunhai Tong, Xiangtai Li, Xuelong Li, et~al.
\newblock Muddit: Liberating generation beyond text-to-image with a unified discrete diffusion model.
\newblock {\em arXiv preprint arXiv:2505.23606}, 2025.

\bibitem{lmfusion}
Weijia Shi, Xiaochuang Han, Chunting Zhou, Weixin Liang, Xi~Victoria Lin, Luke Zettlemoyer, and Lili Yu.
\newblock Lmfusion: Adapting pretrained language models for multimodal generation.
\newblock {\em arXiv preprint arXiv: 2412.15188}, 2024.

\bibitem{dualtoken}
Wei Song, Yuran Wang, Zijia Song, Yadong Li, Haoze Sun, Weipeng Chen, Zenan Zhou, Jianhua Xu, Jiaqi Wang, and Kaicheng Yu.
\newblock Dualtoken: Towards unifying visual understanding and generation with dual visual vocabularies.
\newblock {\em arXiv preprint arXiv:2503.14324}, 2025.

\bibitem{llamagen}
Peize Sun, Yi Jiang, Shoufa Chen, Shilong Zhang, Bingyue Peng, Ping Luo, and Zehuan Yuan.
\newblock Autoregressive model beats diffusion: Llama for scalable image generation.
\newblock {\em arXiv preprint arXiv:2406.06525}, 2024.

\bibitem{swerdlow2025unidisc}
Alexander Swerdlow, Mihir Prabhudesai, Siddharth Gandhi, Deepak Pathak, and Katerina Fragkiadaki.
\newblock Unified multimodal discrete diffusion.
\newblock {\em arXiv preprint arXiv:2503.20853}, 2025.

\bibitem{ugen}
Hongxuan Tang, Hao Liu, and Xinyan Xiao.
\newblock Ugen: Unified autoregressive multimodal model with progressive vocabulary learning.
\newblock {\em arXiv preprint arXiv:2503.21193}, 2025.

\bibitem{codi2}
Zineng Tang, Ziyi Yang, Mahmoud Khademi, Yang Liu, Chenguang Zhu, and Mohit Bansal.
\newblock Codi-2: In-context, interleaved, and interactive any-to-any generation.
\newblock 2023.

\bibitem{CoDI}
Zineng Tang, Ziyi Yang, Chenguang Zhu, Michael Zeng, and Mohit Bansal.
\newblock Any-to-any generation via composable diffusion.
\newblock {\em NeurIPS}, 36, 2024.

\bibitem{team2024chameleon}
Chameleon Team.
\newblock Chameleon: Mixed-modal early-fusion foundation models.
\newblock {\em arXiv preprint arXiv:2405.09818}, 2024.

\bibitem{team2023gemini}
Gemini Team, Rohan Anil, Sebastian Borgeaud, Yonghui Wu, Jean-Baptiste Alayrac, Jiahui Yu, Radu Soricut, Johan Schalkwyk, Andrew~M Dai, Anja Hauth, et~al.
\newblock Gemini: a family of highly capable multimodal models.
\newblock {\em arXiv preprint arXiv:2312.11805}, 2023.

\bibitem{tian2025unigen}
Rui Tian, Mingfei Gao, Mingze Xu, Jiaming Hu, Jiasen Lu, Zuxuan Wu, Yinfei Yang, and Afshin Dehghan.
\newblock Unigen: Enhanced training \& test-time strategies for unified multimodal understanding and generation.
\newblock {\em arXiv preprint arXiv:2505.14682}, 2025.

\bibitem{tong2024cambrian1}
Shengbang Tong, Ellis Brown, Penghao Wu, Sanghyun Woo, Manoj Middepogu, Sai~Charitha Akula, Jihan Yang, Shusheng Yang, Adithya Iyer, Xichen Pan, Austin Wang, Rob Fergus, Yann LeCun, and Saining Xie.
\newblock Cambrian-1: A fully open, vision-centric exploration of multimodal llms, 2024.

\bibitem{tong2024metamorph}
Shengbang Tong, David Fan, Jiachen Zhu, Yunyang Xiong, Xinlei Chen, Koustuv Sinha, Michael Rabbat, Yann LeCun, Saining Xie, and Zhuang Liu.
\newblock Metamorph: Multimodal understanding and generation via instruction tuning.
\newblock {\em arXiv preprint arXiv:2412.14164}, 2024.

\bibitem{llama}
Hugo Touvron, Thibaut Lavril, Gautier Izacard, Xavier Martinet, Marie{-}Anne Lachaux, Timoth{\'{e}}e Lacroix, Baptiste Rozi{\`{e}}re, Naman Goyal, Eric Hambro, Faisal Azhar, Aur{\'{e}}lien Rodriguez, Armand Joulin, Edouard Grave, and Guillaume Lample.
\newblock Llama: Open and efficient foundation language models.
\newblock {\em CoRR}, abs/2302.13971, 2023.

\bibitem{wan2025}
Ang Wang, Baole Ai, Bin Wen, Chaojie Mao, Chen-Wei Xie, Di Chen, Feiwu Yu, Haiming Zhao, Jianxiao Yang, Jianyuan Zeng, Jiayu Wang, Jingfeng Zhang, Jingren Zhou, Jinkai Wang, Jixuan Chen, Kai Zhu, Kang Zhao, Keyu Yan, Lianghua Huang, Mengyang Feng, Ningyi Zhang, Pandeng Li, Pingyu Wu, Ruihang Chu, Ruili Feng, Shiwei Zhang, Siyang Sun, Tao Fang, Tianxing Wang, Tianyi Gui, Tingyu Weng, Tong Shen, Wei Lin, Wei Wang, Wei Wang, Wenmeng Zhou, Wente Wang, Wenting Shen, Wenyuan Yu, Xianzhong Shi, Xiaoming Huang, Xin Xu, Yan Kou, Yangyu Lv, Yifei Li, Yijing Liu, Yiming Wang, Yingya Zhang, Yitong Huang, Yong Li, You Wu, Yu Liu, Yulin Pan, Yun Zheng, Yuntao Hong, Yupeng Shi, Yutong Feng, Zeyinzi Jiang, Zhen Han, Zhi-Fan Wu, and Ziyu Liu.
\newblock Wan: Open and advanced large-scale video generative models.
\newblock {\em arXiv preprint arXiv:2503.20314}, 2025.

\bibitem{textatlas5m}
Alex~Jinpeng Wang, Dongxing Mao, Jiawei Zhang, Weiming Han, Zhuobai Dong, Linjie Li, Yiqi Lin, Zhengyuan Yang, Libo Qin, Fuwei Zhang, et~al.
\newblock Textatlas5m: A large-scale dataset for dense text image generation.
\newblock {\em arXiv preprint arXiv:2502.07870}, 2025.

\bibitem{selftok}
Bohan Wang, Zhongqi Yue, Fengda Zhang, Shuo Chen, Li'an Bi, Junzhe Zhang, Xue Song, Kennard~Yanting Chan, Jiachun Pan, Weijia Wu, Mingze Zhou, Wang Lin, Kaihang Pan, Saining Zhang, Liyu Jia, Wentao Hu, Wei Zhao, and Hanwang Zhang.
\newblock Discrete visual tokens of autoregression, by diffusion, and for reasoning.
\newblock 2025.

\bibitem{ILLUME}
Chunwei Wang, Guansong Lu, Junwei Yang, Runhui Huang, Jianhua Han, Lu Hou, Wei Zhang, and Hang Xu.
\newblock {ILLUME:} illuminating your llms to see, draw, and self-enhance.
\newblock {\em arXiv preprint arXiv:2412.06673}, 2024.

\bibitem{wang2025fudokidiscreteflowbasedunified}
Jin Wang, Yao Lai, Aoxue Li, Shifeng Zhang, Jiacheng Sun, Ning Kang, Chengyue Wu, Zhenguo Li, and Ping Luo.
\newblock Fudoki: Discrete flow-based unified understanding and generation via kinetic-optimal velocities.
\newblock 2025.

\bibitem{wang2023modelscope}
Jiuniu Wang, Hangjie Yuan, Dayou Chen, Yingya Zhang, Xiang Wang, and Shiwei Zhang.
\newblock Modelscope text-to-video technical report.
\newblock {\em arXiv preprint arXiv:2308.06571}, 2023.

\bibitem{wang2024emu3}
Xinlong Wang, Xiaosong Zhang, Zhengxiong Luo, Quan Sun, Yufeng Cui, Jinsheng Wang, Fan Zhang, Yueze Wang, Zhen Li, Qiying Yu, et~al.
\newblock Emu3: Next-token prediction is all you need.
\newblock {\em arXiv preprint arXiv:2409.18869}, 2024.

\bibitem{wang2023lavie}
Yaohui Wang, Xinyuan Chen, Xin Ma, Shangchen Zhou, Ziqi Huang, Yi Wang, Ceyuan Yang, Yinan He, Jiashuo Yu, Peiqing Yang, et~al.
\newblock Lavie: High-quality video generation with cascaded latent diffusion models.
\newblock {\em IJCV}, 2024.

\bibitem{mio}
Zekun Wang, King Zhu, Chunpu Xu, Wangchunshu Zhou, Jiaheng Liu, Yibo Zhang, Jiashuo Wang, Ning Shi, Siyu Li, Yizhi Li, Haoran Que, Zhaoxiang Zhang, Yuanxing Zhang, Ge Zhang, Ke Xu, Jie Fu, and Wenhao Huang.
\newblock Mio: A foundation model on multimodal tokens.
\newblock {\em arXiv preprint arXiv: 2409.17692}, 2024.

\bibitem{wu2025omnigen2}
Chenyuan Wu, Pengfei Zheng, Ruiran Yan, Shitao Xiao, Xin Luo, Yueze Wang, Wanli Li, Xiyan Jiang, Yexin Liu, Junjie Zhou, Ze Liu, Ziyi Xia, Chaofan Li, Haoge Deng, Jiahao Wang, Kun Luo, Bo Zhang, Defu Lian, Xinlong Wang, Zhongyuan Wang, Tiejun Huang, and Zheng Liu.
\newblock Omnigen2: Exploration to advanced multimodal generation.
\newblock {\em arXiv preprint arXiv:2506.18871}, 2025.

\bibitem{liquid}
Junfeng Wu, Yi Jiang, Chuofan Ma, Yuliang Liu, Hengshuang Zhao, Zehuan Yuan, Song Bai, and Xiang Bai.
\newblock Liquid: Language models are scalable multi-modal generators.
\newblock {\em arXiv preprint arXiv:2412.04332}, 2024.

\bibitem{wu2023tune}
Jay~Zhangjie Wu, Yixiao Ge, Xintao Wang, Stan~Weixian Lei, Yuchao Gu, Yufei Shi, Wynne Hsu, Ying Shan, Xiaohu Qie, and Mike~Zheng Shou.
\newblock Tune-a-video: One-shot tuning of image diffusion models for text-to-video generation.
\newblock In {\em ICCV}, 2023.

\bibitem{wu2023next}
Shengqiong Wu, Hao Fei, Leigang Qu, Wei Ji, and Tat-Seng Chua.
\newblock Next-gpt: Any-to-any multimodal llm.
\newblock {\em arXiv preprint arXiv:2309.05519}, 2023.

\bibitem{wu2025openuni}
Size Wu, Zhonghua Wu, Zerui Gong, Qingyi Tao, Sheng Jin, Qinyue Li, Wei Li, and Chen~Change Loy.
\newblock Openuni: A simple baseline for unified multimodal understanding and generation.
\newblock 2025.

\bibitem{wu2025harmon}
Size Wu, Wenwei Zhang, Lumin Xu, Sheng Jin, Zhonghua Wu, Qingyi Tao, Wentao Liu, Wei Li, and Chen~Change Loy.
\newblock Harmonizing visual representations for unified multimodal understanding and generation, 2025.

\bibitem{vila-u}
Yecheng Wu, Zhuoyang Zhang, Junyu Chen, Haotian Tang, Dacheng Li, Yunhao Fang, Ligeng Zhu, Enze Xie, Hongxu Yin, Li Yi, et~al.
\newblock Vila-u: a unified foundation model integrating visual understanding and generation.
\newblock {\em arXiv preprint arXiv:2409.04429}, 2024.

\bibitem{xiao2025haploomni}
Yicheng Xiao, Lin Song, Rui Yang, Cheng Cheng, Zunnan Xu, Zhaoyang Zhang, Yixiao Ge, Xiu Li, and Ying Shan.
\newblock Haploomni: Unified single transformer for multimodal video understanding and generation.
\newblock {\em arXiv preprint arXiv:2506.02975}, 2025.

\bibitem{sana}
Enze Xie, Junsong Chen, Yuyang Zhao, Jincheng Yu, Ligeng Zhu, Yujun Lin, Zhekai Zhang, Muyang Li, Junyu Chen, Han Cai, et~al.
\newblock Sana 1.5: Efficient scaling of training-time and inference-time compute in linear diffusion transformer, 2025.

\bibitem{xie2025sana}
Enze Xie, Junsong Chen, Yuyang Zhao, Jincheng Yu, Ligeng Zhu, Yujun Lin, Zhekai Zhang, Muyang Li, Junyu Chen, Han Cai, et~al.
\newblock Sana 1.5: Efficient scaling of training-time and inference-time compute in linear diffusion transformer, 2025.

\bibitem{Xie_2023_ICCV}
Jinheng Xie, Yuexiang Li, Yawen Huang, Haozhe Liu, Wentian Zhang, Yefeng Zheng, and Mike~Zheng Shou.
\newblock Boxdiff: Text-to-image synthesis with training-free box-constrained diffusion.
\newblock In {\em ICCV}, pages 7452--7461, 2023.

\bibitem{showo}
Jinheng Xie, Weijia Mao, Zechen Bai, David~Junhao Zhang, Weihao Wang, Kevin~Qinghong Lin, Yuchao Gu, Zhijie Chen, Zhenheng Yang, and Mike~Zheng Shou.
\newblock Show-o: One single transformer to unify multimodal understanding and generation.
\newblock In {\em ICLR}, 2025.

\bibitem{MUSE-VL}
Rongchang Xie, Chen Du, Ping Song, and Chang Liu.
\newblock {MUSE-VL:} modeling unified {VLM} through semantic discrete encoding.
\newblock {\em arXiv preprint arXiv:2411.17762}, 2024.

\bibitem{xing2023dynamicrafter}
Jinbo Xing, Menghan Xia, Yong Zhang, Haoxin Chen, Xintao Wang, Tien-Tsin Wong, and Ying Shan.
\newblock Dynamicrafter: Animating open-domain images with video diffusion priors.
\newblock {\em arXiv preprint arXiv:2310.12190}, 2023.

\bibitem{xu2024pllava}
Lin Xu, Yilin Zhao, Daquan Zhou, Zhijie Lin, See~Kiong Ng, and Jiashi Feng.
\newblock Pllava: Parameter-free llava extension from images to videos for video dense captioning.
\newblock {\em arXiv preprint arXiv:2404.16994}, 2024.

\bibitem{xu2025pisces}
Zhiyang Xu, Jiuhai Chen, Zhaojiang Lin, Xichen Pan, Lifu Huang, Tianyi Zhou, Madian Khabsa, Qifan Wang, Di Jin, Michihiro Yasunaga, et~al.
\newblock Pisces: An auto-regressive foundation model for image understanding and generation.
\newblock {\em arXiv preprint arXiv:2506.10395}, 2025.

\bibitem{qwen2.5}
An Yang, Baosong Yang, Beichen Zhang, Binyuan Hui, Bo Zheng, Bowen Yu, Chengyuan Li, Dayiheng Liu, Fei Huang, Haoran Wei, Huan Lin, Jian Yang, Jianhong Tu, Jianwei Zhang, Jianxin Yang, Jiaxi Yang, Jingren Zhou, Junyang Lin, Kai Dang, Keming Lu, Keqin Bao, Kexin Yang, Le Yu, Mei Li, Mingfeng Xue, Pei Zhang, Qin Zhu, Rui Men, Runji Lin, Tianhao Li, Tingyu Xia, Xingzhang Ren, Xuancheng Ren, Yang Fan, Yang Su, Yichang Zhang, Yu Wan, Yuqiong Liu, Zeyu Cui, Zhenru Zhang, and Zihan Qiu.
\newblock Qwen2.5 technical report.
\newblock {\em arXiv preprint arXiv:2412.15115}, 2024.

\bibitem{MMAR}
Jian Yang, Dacheng Yin, Yizhou Zhou, Fengyun Rao, Wei Zhai, Yang Cao, and Zheng{-}Jun Zha.
\newblock {MMAR:} towards lossless multi-modal auto-regressive probabilistic modeling.
\newblock {\em arXiv preprint arXiv:2410.10798}, 2024.

\bibitem{yang2025mmada}
Ling Yang, Ye Tian, Bowen Li, Xinchen Zhang, Ke Shen, Yunhai Tong, and Mengdi Wang.
\newblock Mmada: Multimodal large diffusion language models.
\newblock {\em arXiv preprint arXiv:2505.15809}, 2025.

\bibitem{yang2024cogvideox}
Zhuoyi Yang, Jiayan Teng, Wendi Zheng, Ming Ding, Shiyu Huang, Jiazheng Xu, Yuanming Yang, Wenyi Hong, Xiaohan Zhang, Guanyu Feng, et~al.
\newblock Cogvideox: Text-to-video diffusion models with an expert transformer.
\newblock {\em arXiv preprint arXiv:2408.06072}, 2024.

\bibitem{mmmu}
Xiang Yue, Yuansheng Ni, Tianyu Zheng, Kai Zhang, Ruoqi Liu, Ge Zhang, Samuel Stevens, Dongfu Jiang, Weiming Ren, Yuxuan Sun, Cong Wei, Botao Yu, Ruibin Yuan, Renliang Sun, Ming Yin, Boyuan Zheng, Zhenzhu Yang, Yibo Liu, Wenhao Huang, Huan Sun, Yu Su, and Wenhu Chen.
\newblock {MMMU:} {A} massive multi-discipline multimodal understanding and reasoning benchmark for expert {AGI}.
\newblock In {\em {CVPR}}, pages 9556--9567. {IEEE}, 2024.

\bibitem{siglip}
Xiaohua Zhai, Basil Mustafa, Alexander Kolesnikov, and Lucas Beyer.
\newblock Sigmoid loss for language image pre-training, 2023.

\bibitem{rmsnorm}
Biao Zhang and Rico Sennrich.
\newblock Root mean square layer normalization.
\newblock In {\em NeurIPS}, 2019.

\bibitem{show1}
David~Junhao Zhang, Jay~Zhangjie Wu, Jia-Wei Liu, Rui Zhao, Lingmin Ran, Yuchao Gu, Difei Gao, and Mike~Zheng Shou.
\newblock Show-1: Marrying pixel and latent diffusion models for text-to-video generation.
\newblock {\em arXiv preprint arXiv:2309.15818}, 2023.

\bibitem{Nexus-Gen}
Hong Zhang, Zhongjie Duan, Xingjun Wang, Yingda Chen, Yuze Zhao, and Yu Zhang.
\newblock Nexus-gen: A unified model for image understanding, generation, and editing.
\newblock {\em arXiv preprint arXiv:2504.21356}, 2025.

\bibitem{zhang2024internlm}
Pan Zhang, Xiaoyi Dong, Yuhang Zang, Yuhang Cao, Rui Qian, Lin Chen, Qipeng Guo, Haodong Duan, Bin Wang, Linke Ouyang, et~al.
\newblock Internlm-xcomposer-2.5: A versatile large vision language model supporting long-contextual input and output.
\newblock {\em arXiv preprint arXiv:2407.03320}, 2024.

\bibitem{zhang2024long}
Peiyuan Zhang, Kaichen Zhang, Bo Li, Guangtao Zeng, Jingkang Yang, Yuanhan Zhang, Ziyue Wang, Haoran Tan, Chunyuan Li, and Ziwei Liu.
\newblock Long context transfer from language to vision.
\newblock {\em arXiv preprint arXiv:2406.16852}, 2024.

\bibitem{zhang2023i2vgen}
Shiwei Zhang, Jiayu Wang, Yingya Zhang, Kang Zhao, Hangjie Yuan, Zhiwu Qin, Xiang Wang, Deli Zhao, and Jingren Zhou.
\newblock I2vgen-xl: High-quality image-to-video synthesis via cascaded diffusion models.
\newblock {\em arXiv preprint arXiv:2311.04145}, 2023.

\bibitem{llava-next-video}
Yuanhan Zhang, Jinming Wu, Wei Li, Bo Li, Zejun Ma, Ziwei Liu, and Chunyuan Li.
\newblock Video instruction tuning with synthetic data.
\newblock {\em arXiv preprint arXiv:2410.02713}, 2024.

\bibitem{monoformer}
Chuyang Zhao, Yuxing Song, Wenhao Wang, Haocheng Feng, Errui Ding, Yifan Sun, Xinyan Xiao, and Jingdong Wang.
\newblock Monoformer: One transformer for both diffusion and autoregression.
\newblock {\em arXiv preprint arXiv:2409.16280}, 2024.

\bibitem{zhou2025transfusion}
Chunting Zhou, LILI YU, Arun Babu, Kushal Tirumala, Michihiro Yasunaga, Leonid Shamis, Jacob Kahn, Xuezhe Ma, Luke Zettlemoyer, and Omer Levy.
\newblock Transfusion: Predict the next token and diffuse images with one multi-modal model.
\newblock In {\em ICLR}, 2025.

\bibitem{vargpt}
Xianwei Zhuang, Yuxin Xie, Yufan Deng, Liming Liang, Jinghan Ru, Yuguo Yin, and Yuexian Zou.
\newblock Vargpt: Unified understanding and generation in a visual autoregressive multimodal large language model, 2025.

\bibitem{OmniMamba}
Jialv Zou, Bencheng Liao, Qian Zhang, Wenyu Liu, and Xinggang Wang.
\newblock Omnimamba: Efficient and unified multimodal understanding and generation via state space models, 2025.

\end{thebibliography}
    }

\end{document}